\documentclass[mnsc,nonblindrev]{informs3_hide}
\OneAndAHalfSpacedXI 

\usepackage[english]{babel}
\usepackage[autostyle, english = american]{csquotes}
\MakeOuterQuote{"}
\usepackage[colorlinks,linkcolor=blue, citecolor=blue]{hyperref}
\usepackage[normalem]{ulem}

\usepackage{cleveref}
\crefname{subsection}{section}{subsections}
\usepackage{eqnarray}

\RequirePackage{algorithm}
\RequirePackage{algorithmic}
\usepackage{amsmath, bbm, enumitem, xparse, bm, graphicx, lipsum}
\usepackage{amssymb}
\usepackage{subcaption}

\newcommand{\eps}{\varepsilon}
\newcommand{\bI}{\mathbbm{1}}

\newcommand{\ALG}{\mathsf{ALG}}
\newcommand{\ALGD}{\mathsf{ALG}_{\mathsf{Dual}}}
\newcommand{\OPT}{\mathsf{OPT}}
\newcommand{\hOPT}{\hat{\mathsf{OPT}}}
\newcommand{\OPTInner}{\mathsf{OPT}}
\newcommand{\OPTOuter}{\mathsf{OPT}^{\mathsf{Covering}}}
\newcommand{\LInner}{\mathsf{L}}
\newcommand{\LOuter}{\mathsf{L}^{\mathsf{Covering}}}
\newcommand{\bLInner}{\bar{L}}
\newcommand{\bLOuter}{\bar{L}^{\mathsf{Covering}}}
\newcommand{\hLInner}{\hat{L}}

\newcommand{\Reg}{\mathsf{Regret}}

\newcommand{\bc}{\bm{c}}
\newcommand{\hbc}{\hat{\bm{c}}}

\newcommand{\tbc}{\tilde{\bm{c}}}

\newcommand{\bx}{\bm{x}}
\newcommand{\tbx}{\tilde{\bm{x}}}
\newcommand{\hbx}{\hat{\bm{x}}}

\usepackage[dvipsnames]{xcolor}

\usepackage{xparse}

\NewDocumentEnvironment{myproof}{o}
{\IfNoValueTF{#1}{\paragraph{{Proof.} }} {\paragraph{{#1.} }} }
{\hfill$\Halmos$}

\usepackage{natbib,bbm,multirow,multicol}
 \bibpunct[, ]{(}{)}{,}{a}{}{,}%
 \def\bibfont{\small}%
\TheoremsNumberedThrough     
\ECRepeatTheorems

\EquationsNumberedThrough    

\begin{document}




\TITLE{
\Large Constrained Online Two-stage Stochastic Optimization:\\
           Algorithm with (and without) Predictions
}

\ARTICLEAUTHORS{%
\AUTHOR{$\text{Piao Hu}^{\dagger}$, $\text{Jiashuo Jiang}^{\dagger}$, $\text{Guodong Lyu}^{\dagger}$, $\text{Hao Su}^{\dagger}$}

\AFF{\  \\
$\dagger~$Hong Kong University of Science and Technology
}
}

\ABSTRACT{
We consider an online two-stage stochastic optimization with long-term constraints over a finite horizon of $T$ periods. At each period, we take the first-stage action, observe a model parameter realization and then take the second-stage action from a feasible set that depends both on the first-stage decision and the model parameter. We aim to minimize the cumulative objective value while guaranteeing that the long-term average second-stage decision belongs to a set. We develop online algorithms for the online two-stage problem from adversarial learning algorithms. Also, the regret bound of our algorithm cam be reduced to the regret bound of embedded adversarial learning algorithms. Based on this framework, we obtain new results under various settings. When the model parameters are drawn from unknown non-stationary distributions and we are given machine-learned predictions of the distributions, we develop a new algorithm from our framework with a regret $O(W_T+\sqrt{T})$, where $W_T$ measures the total inaccuracy of the machine-learned predictions. We then develop another algorithm that works when no machine-learned predictions are given and show the performances.
}



\maketitle

\section{Introduction}\label{sec:intro}
Stochastic optimization is widely used to model the decision making problem with uncertain model parameters. In general, stochastic optimization aims to solve the problem with formulation $\min_{\bc\in\mathcal{C}} \mathbb{E}_{\theta}[F_{\theta}(\bc)]$, where $\theta$ models parameter uncertainty and we optimize the objective on average. An important class of stochastic optimization models is the \textit{two-stage model}, where the problem is further divided into two stages. In particular, at the first stage, we decide the first-stage decision $\bc$ without knowing the exact value of $\theta$. At the second stage, after a realization of the uncertain data becomes known, an optimal second stage decision $\bx$ is made by solving an optimization problem parameterized by both $\bc$ and $\theta$, where one of the constraint can be formulated as $x\in\mathcal{B}(c, \theta)$. Here, $F_{\theta}(\bc)$ denotes the optimal objective value of the second-stage optimization problem. Two-stage stochastic optimization has numerous applications, including  transportation, logistics, financial instruments, and supply chain, among others \citep{birge2011introduction}.

In this paper, we focus on an ``online'' extension of the classical two-stage stochastic optimization over a finite horizon of $T$ periods. Subsequently, at each period $t$, we first decide the first-stage decision $\bc_t\in\mathcal{C}$, then observe the value of model parameter $\theta_t$, which is assumed to be drawn from an \textit{unknown} distribution $P_t$, and finally decide the second-stage decision $\bx_t$. In addition to requiring $\bx_t$ belonging to a constraint set parameterized by $\bc_t$ and $\theta_t$, we also need to satisfy a long-term global constraint of the following form: $\frac{1}{T}\cdot\sum_{t=1}^{T}\bm{g}_{\theta_t}(\bc_t, \bm{x}_t)\in\mathcal{B}$. We aim to optimize the total objective value over the entire horizon, and we measure the performance of our online policy by ``regret'', the additive loss of the online policy compared to the \textit{optimal dynamic policy} which is aware of $P_t$ for each period $t$.

Indeed, one motivating example for our model is from supply chain management and our model has many other applications. Usually, the supply chain system of a retail company is composed of two layers. One is the centralized \textit{warehouses} layer, and the other is a local \textit{retail stores} layer. At each period, the company needs to decide how much inventory to be invested into the warehouses, which is the first-stage deicion $\bc$, and then, after the customer demand realizes, the company needs to decide how to transfer the inventory from each warehouse to the downstream retail stores to fulfill customer demand, which is the second-stage decision $\bx$. It is common in practice that over the entire horizon, the total inventory received at each retail store cannot surpass certain thresholds due to some capacity constraints, or the service level of the customers for each retail store (defined as the total number of times customer demand is fully fulfilled or the fraction of fulfilled customer demand over total demand, for each retail store) cannot be smaller than certain thresholds. These operational requirements induce long-term constraints into our model.

\subsection{Main Results}\label{sec:OurResults}

Our main result is an online algorithm to achieve a sublinear regret for the two-stage model with non-stationary distributions, i.e. $P_t$ is non-homogeneous over $t\in[T]$. The distribution $P_t$ is assumed to be unknown for each $t\in[T]$, however, in practice, historical data are usually available for the distribution $P_t$, from which we can use some machine learning methods to form a prediction for $P_t$. Therefore, we consider a prediction setting where we have a machine-learned prediction $\hat{P}_t$ of $P_t$, for each $t\in[T]$, at the beginning of the horizon. Note that the prediction $\hat{P}_t$ can be different from the true distribution $P_t$ due to certain prediction error. We aim to develop online algorithm to incorporate these predictions to guide our decision making under the non-stationary environment, while guaranteeing that the performance of our algorithm is robust to the prediction errors. To this end, we first introduce parameter $W$ to capture the total inaccuracy of the pre-given predictions. We show that no online algorithm can achieve a regret bound better than $\Omega(W)$ compared to the offline optimum. We then develop an algorithm to match this lower bound.
we introduce a dual variable for each long-term constraint and we update the dual variable at each period to control how the budget of each long-term constraint is consumed. Given the fixed dual variable, we can formulate a two-stage stochastic optimization problem from the Lagrangian dual function, and the first and second stage decision at each period can be obtained from solving this two-stage stochastic optimization problem. An adversarial learning algorithm is finally utilized to update the dual variable, based on the feedback from the solution of this two-stage stochastic optimization problem. The two-stage problem is formulated differently for different periods to handle the non-stationary of the underlying distributions. The formulation relies on the predictions, and as a result, our algorithm, which we name \textit{Informative Adversarial Learning} (IAL) algorithm, naturally combines the information provided by the predictions into the adversarial learning algorithm of the dual variables for the long-term constraints. Our IAL algorithm achieves a regret bound $O(W+\sqrt{T})$, which matches the lower bound $\Omega(W)$.

We then extend to the setting where the predictions are absent and one has to learn the current distribution purely from the past observations. Note that if we allow the underlying distributions to be arbitrarily non-stationary (this will become the adversarial setting), no sublinear regret can be obtained. We consider an intermediate setting between stationary setting and arbitrary non-stationary setting. To be specific, we assume the underlying distributions to be stationary but we allow adversarial corruptions to the realizations. There can be in total $W$ number of corruptions. We modify the IAL algorithm in that we uses another adversarial learning algorithm such as online gradient descent to update the first-stage decision, thus handling the influence of the adversarial corruptions. Therefore, we adopt two adversarial learning algorithms to update the first-stage decision and the dual variable simultaneously, while the second-stage decision is determined afterwards. In this sense, we name our algorithm \textit{Doubly Adversarial Learning} (DAL) algorithm and we show that it achieves a regret bound of $O(W+\sqrt{T})$, which also matches the lower bound $\Omega(W)$ in terms of the dependency on the number of corruptions.

\subsection{Related Work}
\textbf{Algorithm with predictions}. There has been a recent trend on studying the performances of online algorithms with pre-existing machine-learned predictions. In this trend, the competitive ratio bound is a popular metric to measure the performance of the algorithm, which is defined as $\text{ALG}/\text{OPT}$, where $\text{ALG}$ stands for the reward of the algorithm and $\text{OPT}$ stands for the optimality benchmark. Competitive ratio bound has been investigated in a series of papers (e.g. \citet{mahdian2012online, antoniadis2020secretary, balseiro2022single, banerjee2022online, jin2022online, golrezaei2023online}). Though the competitive ratio is a robust measure and can guarantee the performance of the algorithm even when the predictions are totally inaccurate or absent, see for example \cite{immorlica2019adversarial, castiglioni2022online, balseiro2023best}, it transfers to a regret bound that is linear in $T$. The additive bound, i.e. the regret bound, for algorithm with predictions have also been studied recently, for example, in the papers \cite{munoz2017revenue, an2023newsvendor, hao2023leveraging}. However, the previous papers focus on the bandit setting without long-term constraints, while our paper considers a general two-stage model with long-term constraints and sublinear regret bounds are derived that incorporates the prediction errors. Note that the non-stationary bandits with knapsack problem has been considered in \citet{lyu2023bandits} with deterministic predictions (rather than a distributional prediction in our setting) and algorithmic design. Our model covers the bandit with knapsack model when the second-stage decision is de-activated. The algorithm with prediction setting has also been considered in other problems such as caching \cite{lykouris2021competitive, rohatgi2020near}, online scheduling \cite{lattanzi2020online}, and the secretary problem \cite{dutting2021secretaries}.

\noindent\textbf{Non-stationary online optimization.} Though the classical online convex optimization allows adversarial input, it considers static benchmark by restricting the benchmark to take a common decision at each period. In order to obtain a sublinear \textit{dynamic regret}, we have to bound the extent of non-stationarity even for the most fundamental multi-arm-bandit problem where the long-term constraint is absent. Various papers have considered the non-stationary online optimization problem with dynamic benchmark, under different measures of non-stationarity (e.g. \citet{zinkevich2003online, besbes2014stochastic, besbes2015non, cheung2020reinforcement, celli2022best}). The papers that most close to ours are \cite{balseiro2023best} and \cite{jiang2020online}, where similar measures of non-stationarity are considered. However, we consider an online two-stage problem, which is more general and the algorithms are different.

\noindent\textbf{Connections with other problems.} Our online model is a natural synthesis of several widely studied models in the existing literature. Roughly speaking, we classify previous models on online learning/optimization into the following two categories: the \textit{bandits-based} model and the \textit{type-based} model. For the bandits-based model, we make the decision and then observe the (possibly stochastic) outcome, which can be adversarially chosen. The representative problems include multi-arm-bandits (MAB) problem, and the more general online convex optimization (OCO) problem. Captured in our model, the second-stage decision is de-activated. For the type-based model, at each period, we first observe the type of the arrival, and we are clear of the possible outcome for each action (which can be type-dependent), and then we decide the action without knowing the type of future periods. Note that in the type-based model, we usually have a global constraint such that the cumulative decision over the entire horizon belongs to a set (otherwise the problem becomes trivial, just select the myopic optimal action at each period), which corresponds to the long-term constraint in our model. The representative problems for the type-based model include online allocation problem and a special case online packing problem where the objective function is linear and $\mathcal{B}(\bm{C}, \bm{\theta})$ is a polyhedron. Captured in our model, the first-stage decision is de-activated. We review the literature on these problems in \Cref{sec:RelatedWork}.

\section{Problem Formulation}\label{sec:formulation}
We consider the online two-stage stochastic optimization problem with long-term constraints in the following general formulation. There is a finite horizon of $T$ periods and at each period $t$, the following events happen in sequence:
\begin{itemize}
  \item[1] we decide the first-stage decision $\bm{c}_t\in\mathcal{C}$ and incur a cost $p(\bm{c}_t)$;
  \item[2] the type $\theta_t$ is drawn independently from an \textit{unknown} distribution $P_t$, and the second-stage objective function $f_{\theta_t}(\cdot)$ and the constraint function $\bm{g}_{\theta_t}(\cdot)=(g_{i,\theta_t}(\cdot))_{i=1}^m$ become known.
  \item[3] we decide the second-stage decision $\bm{x}_t\in\mathcal{K}(\theta_t, \bm{c}_t)$, where $\mathcal{K}(\theta_t, \bm{c}_t)$ is a feasible set parameterized by both the type $\theta_t$ and the first-stage decision $\bm{c}_t$, and incur an objective value $f_{\theta_t}(\cdot)$.
\end{itemize}
At the end of the entire horizon, the long-term constraints $\frac{1}{T}\cdot\sum_{t=1}^{T}\bx_t\in\mathcal{B}(\bm{C}, \bm{\theta})$ need to be satisfied, which is characterized as follows, following \cite{mahdavi2012trading},
\begin{equation}\label{eqn:LongConstraint}
  \frac{1}{T}\cdot\sum_{t=1}^{T}\bm{g}_{\theta_t}(\bc_t, \bm{x}_t)\leq\bm{\beta},
\end{equation}
with $\bm{\beta}\in(0,1)^m$. We aim to minimize the objective
\begin{equation}\label{eqn:LongObjective}
\sum_{t=1}^{T}\left(p(\bm{c}_t)+f_{\theta_t}(\bm{x}_t)\right).
\end{equation}
Any online policy is feasible as long as $\bm{c}_t$ and $\bm{x}_t$ are \textit{agnostic} to the future realizations while satisfying \eqref{eqn:LongConstraint}. The benchmark is the \textit{dynamic optimal policy}, denoted by $\pi^*$, who is aware of the distributions $\bm{P}=(P_t)_{t=1}^T$ but still the decisions $\bm{c}_t$ and $\bm{x}_t$ have to be agnostic to future realizations. Note that this benchmark is more power than the optimal online policy we are seeking for who is unaware of the distributions $\bm{P}$, and the optimal policy can be dynamic.
We are interested in developing a feasible online policy $\pi$ with known \textit{regret} upper bound compared to the optimal policy:
\begin{equation}\label{def:regret}
  \Reg(\pi, T)=\mathbb{E}_{\bm{\theta}\sim\bm{P}}\left[\ALG(\pi,\bm{\theta})-\ALG(\pi^*, \bm{\theta})\right],
\end{equation}
where $\ALG(\pi,\bm{\theta})$ denotes the objective value of policy $\pi$ on the sequence $\bm{\theta}$. The optimal policy can possess very complicated structures thus lacks tractability. Therefore, we develop a tractable lower bound of $\mathbb{E}_{\bm{\theta}\sim\bm{P}}[\ALG(\pi^*, \bm{\theta})]$. The lower bound is given by the optimization problem below.
\begin{align}
 \tag{OPT}\OPT= \min & ~~\sum_{t=1}^{T}\mathbb{E}_{\tbc_t, \tbx_t, \theta_t}\left[p(\tilde{\bm{c}}_t)+f_{\theta_t}(\tilde{\bm{x}}_t)\right], \label{lp:OPT}\\
  \mbox{s.t.} & ~~\frac{1}{T}\cdot\sum_{t=1}^{T}\mathbb{E}_{\tbc_t, \tbx_t, \theta_t}[\bm{g}_{\theta_t}(\tbc_t, \tbx_t)]\leq\bm{\beta}, \nonumber\\
  & ~~\tbx_t\in\mathcal{K}(\theta_t, \tbc_t), \tbc_t\in\mathcal{C}, \forall t.\nonumber
\end{align}
Here, $\tbc_t$ and $\tbx_t$ are random variables for each $t\in[T]$ and the distribution of $\tbc_t$ is \textit{independent} of $\theta_t$. Note that we only need the formulation of \eqref{lp:OPT} to conduct our theoretical analysis. We never require to really solve the optimization problem \eqref{lp:OPT}.
Clearly, if we let $\tbc_t$ (resp. $\tbx_t$) denote the \textit{marginal distribution} of the first-stage (resp. second-stage) decision made by the optimal policy, the we would have a feasible solution to \Cref{lp:OPT} while the objective value equals $\mathbb{E}_{\bm{\theta}\sim\bm{P}}[\ALG(\pi^*, \bm{\theta})]$. This argument leads to the following lemma.
\begin{lemma}[forklore]\label{lem:LowerBound}
 $\OPT\leq \mathbb{E}_{\bm{\theta}\sim\bm{P}}[\ALG(\pi^*, \bm{\theta})]$.
\end{lemma}
Throughout the paper, we make the following assumptions.
\begin{assumption}\label{assump:main}
The following conditions are satisfied:
\begin{itemize}
  \item[a.] (convexity) The function $p(\cdot)$ is a convex function with $p(\bm{0})=0$ and $\mathcal{C}$ is a compact and convex set which contains $\bm{0}$. The set $\mathcal{A}\subset[0,1]^m$ is a convex set. Also, the functions $f_{\theta}(\cdot)$ is convex in $\bm{x}$ and $\bm{g}_{\theta}(\cdot, \cdot)$ is convex in both $\bc$ and $\bm{x}$, for any $\theta$.
  \item[b.] (compactness) The set $\mathcal{K}(\theta_t, \bm{c}_t)$ is a polyhedron given by $\mathcal{K}(\theta_t, \bm{c}_t)=\{\bm{x}\in\mathbb{R}^m: \bm{0}\leq\bm{x}\text{~and~}B_{\theta_t}\bm{x}\leq\bm{c}_t\}\cap\mathcal{K}$ where $\mathcal{K}$ is a compact convex set.
  \item[c.] (boundedness) For any $\bm{x}\in\mathcal{K}$ and any $\theta$, we have $f_{\theta}(\bm{x})\in[-1,1]$ and $\bm{g}_{\theta}(\bm{x})\in[0,1]^m$. Moreover, it holds that $f_{\theta}(\bm{0})=0$ and $\bm{g}_{\theta}(\bm{0}, \bm{0})=\bm{0}$.
\end{itemize}
\end{assumption}

The conditions listed in \Cref{assump:main} are standard in the literature, which mainly ensure the convexity of the problem as well as the boundedness of the objective functions and the feasible set. Finally, in condition c, we assume that there always exists $\bm{0}$, which denotes a \textit{null} action that has no influence on the accumulated objective value and the constraints.

The Lagrangian dual problem of \eqref{lp:OPT} is given below:
\begin{align}
&\max_{\bm{\lambda}\geq0}\min_{\tbc_t\in\mathcal{C}}~~ \LInner(\bm{C}, \bm{\lambda})=\sum_{t=1}^{T}\mathbb{E}\left[p(\tbc_t)-\frac{1}{T}\cdot\sum_{i=1}^{m}\lambda_i+\right.  \nonumber\\
&\tag{Dual} \left.\min_{\tbx_t\in\mathcal{K}(\theta_t,\tbc_t)} \left\{f_{\theta_t}(\tbx_t)+\sum_{i=1}^{m}\frac{\lambda_i\cdot g_{i,\theta_t}(\tbc_t, \tbx_t)}{T\cdot\beta_i}\right\}\right].  \label{lp:DualInner}
\end{align}
where $\bm{C}=(\tbc_t)_{t=1}^T$ and we note that the distribution of $\tbc_t$ is independent of $\theta_t$.
From weak duality, \eqref{lp:DualInner} also serves as a lower bound of $\mathbb{E}_{\bm{\theta}\sim\bm{P}}[\ALG(\pi^*, \bm{\theta})]$. Therefore, in what follows, we compare against \eqref{lp:DualInner} to derive our regret bound.

\section{Non-stationary Setting with Predictions}\label{sec:nonstationary}

In this section, we consider the non-stationary setting where $P_t$ is \textit{unknown} and \textit{non-homogeneous} for each $t$. Since we are comparing against the dynamic optimal policy, our definition of regret in \eqref{def:regret} falls into the range of \textit{dynamic regret}. It has been known in the literature (e.g. \cite{besbes2014stochastic}) that without additional restrictions on the true distributions $\bm{P}=(P_t)_{t=1}^T$, one cannot achieve a dynamic regret that is sublinear in $T$. Therefore, we seek for sublinear regret with the help of \textit{additional information}. In practice, the horizon can be repeated for multiple times, from which we obtain some historical dataset over the true distributions $\bm{P}=(P_t)_{t=1}^T$. It is common to utilize some machine learning methods to learn the true distributions $\bm{P}=(P_t)_{t=1}^T$. As a result, in this section, we assume that there exists a prediction $\hat{P}_t$ of the true distribution $P_t$, for each $t\in[T]$. The predictions $\hat{\bm{P}}=(\hat{P}_t)_{t=1}^T$ are given at the beginning of the horizon and can be different from the true distributions $\bm{P}=(P_t)_{t=1}^T$ due to some prediction errors. We explore how to utilize these predictions $\hat{\bm{P}}=(\hat{P}_t)_{t=1}^T$ to make our online decisions.


We first derive the regret lower bound and show how the regret should depend on the possible inaccuracy of the predictions. Following \cite{jiang2020online}, we measure the inaccuracy of $\hat{P}_t$ by Wasserstein distance (we refer interested readers to Section 6.3 of \cite{jiang2020online} for benefits of using Wasserstein distance in online decision making), which is defined as follows
\begin{equation}\label{eqn:Wasserdist}
W(\hat{P}_t, P_t):=\inf_{Q\in\mathcal{F}(\hat{P}_t,P_t)}\int d(\theta, \theta')dQ(\theta, \theta'),
\end{equation}
where $d(\theta, \theta')=\|(f_{\theta}, \bm{g}_{\theta})-(f_{\theta'}, \bm{g}_{\theta'})\|_{\infty}$ and $\mathcal{F}(\hat{P}_t, P_t)$ denotes the set of all joint distributions for $(\theta, \theta')$ with marginal distributions $\hat{P}_t$ and $P_t$. In the following theorem, we show that one cannot break a linear dependency on the total inaccuracy of the predictions.
\begin{theorem}\label{thm:priorLower}
Let $W_T=\sum_{t=1}^{T}W(\hat{P}_t, P_t)$ be the total measure of inaccuracy, where $W(\hat{P}_t, P_t)$ is defined in \eqref{eqn:Wasserdist}. For any feasible online policy $\pi$ that only knows predictions, there always exists $\bm{P}$ and $\hat{\bm{P}}$ such that
\[
\Reg(\pi, T)=\mathbb{E}_{\bm{\theta}\sim\bm{P}}[\ALG(\pi,\bm{\theta})-\ALG(\pi^*,\bm{\theta})]\geq
\Omega\left(W_T\right),
\]
where $\pi^*$ denotes the optimal policy that knows true distributions $\bm{P}$.
\end{theorem}

We then derive our online algorithm with a regret that matches the lower bound established above. When we have a predictions available, an efficient way would be to ``greedily'' select $\bm{c}_t$ with the help of predictions. In order to describe our idea, we denote by $\{\hat{\bm{\lambda}}^*, \{\hbc^*_t\}_{\forall t\in[T]}\}$ one optimal solution to
\begin{align}
&\max_{\bm{\lambda}\geq0}\min_{\bm{c}_t\in\mathcal{C}} \hLInner(\bm{C}, \bm{\lambda})=\mathbb{E}_{\bm{\theta}\sim\hat{\bm{P}}}\left[\sum_{t=1}^{T}\left(p(\bc_t)-\frac{1}{T}\cdot\sum_{i=1}^{m}\lambda_i\right.\right.  \nonumber\\
&\left.\left.+
\min_{\bx_t\in\mathcal{K}(\theta_t,\bc_t)}f_{\theta_t}(\bx_t)+\sum_{i=1}^{m}\frac{\lambda_i\cdot g_{i,\theta_t}(\bc_t, \bx_t)}{T\cdot\beta_i}\right)\right],  \label{eqn:121901}
\end{align}
where $\hat{\bm{P}}=(\hat{P}_t)_{t=1}^T$. Note that the value of \eqref{eqn:121901} is equivalent to the value of \eqref{lp:DualInner} if $\hat{\bm{P}}=\bm{P}$. Then, we define for each $i\in[m]$,
\begin{equation}\label{eqn:121902}
\hat{\beta}_{i,t}:=\mathbb{E}_{\theta\sim\hat{P}_t}\left[ g_{i,\theta}(\hbc^*_t, \hbx^*_t(\theta)) \right],
\end{equation}
with
\[
\hbx^*_t(\theta)\in\text{argmin}_{\bx_t\in\mathcal{K}(\theta,\hbc^*_t)}f_{\theta}(\bx_t)+\sum_{i=1}^{m}\frac{\hat{\lambda}^*_i\cdot g_{i,\theta}(\hbc^*_t, \bx_t)}{T\cdot\beta_i}.
\]
Here, $\hat{\bm{\beta}}_t$ can be interpreted as the \textit{predictions-informed} target levels to achieve, at each period $t$. To be more concrete, if $\hat{P}_t=P_t$ for each $t$, then $\hat{\bm{\beta}}$ is exactly the value of the constraint functions at period $t$ in \eqref{lp:OPT}, which is a good reference to stick to. We then define
\begin{align}
&\hLInner_t(\bm{c}, \bm{\lambda}):=\mathbb{E}_{\theta\sim\hat{P}_t}\left[ p(\bc)-\sum_{i=1}^{m}\frac{\lambda_i\cdot\hat{\beta}_{i,t}}{T\cdot\beta_i}\right. \nonumber\\
&\left.+\min_{\bx\in\mathcal{K}(\theta,\bc)}f_{\theta}(\bx)+\sum_{i=1}^{m}\frac{\lambda_i\cdot g_{i,\theta}(\bc, \bx)}{T\cdot\beta_i} \right]. \label{eqn:121903}
\end{align}
The key ingredient of our analysis is the following.
\begin{lemma}\label{lem:decompose}
It holds that
\begin{equation}\label{eqn:121904}
  \max_{\bm{\lambda}\geq0}\min_{\bm{c}_t\in\mathcal{C}} \hLInner(\bm{C}, \bm{\lambda})=\sum_{t=1}^{T}\max_{\bm{\lambda}\geq0}\min_{\bm{c}_t\in\mathcal{C}} \hLInner_t(\bm{c}_t, \bm{\lambda}),
\end{equation}
and moreover, letting $(\hat{\bm{\lambda}}^*, (\hbc^*_t)_{t=1}^T)$ be the optimal solution to $\max_{\bm{\lambda}\geq0}\min_{\bm{c}_t\in\mathcal{C}} \hLInner(\bm{C}, \bm{\lambda})$ used in the definition \eqref{eqn:121902}, then $(\hat{\bm{\lambda}}^*, \hbc^*_t)$ is an optimal solution to $\max_{\bm{\lambda}\geq0}\min_{\bm{c}_t\in\mathcal{C}} \hLInner_t(\bm{c}_t, \bm{\lambda})$ for each $t\in[T]$. Also, we have
\begin{equation}\label{eqn:121908}
\sum_{i=1}^{m}\sum_{t=1}^{T}\hat{\lambda}^*_{i,t}\hat{\beta}_i=T\cdot\sum_{i=1}^{m}\beta_i\cdot\hat{\lambda}^*_i,
\end{equation}
for each $i\in[m]$.
\end{lemma}
\Cref{lem:decompose} implies that given $\hat{\bm{\lambda}}^*$, we can simply minimize $\hLInner_t(\bm{c}_t, \hat{\bm{\lambda}}^*)$ over $\bm{c}_t$ to get the first-stage decision \footnote{This is a stochastic optimization problem, which can be solved by applying \textit{stochastic gradient descent} over $\bm{c}_t$ or applying \textit{sample average approximation} to get samples of $\theta\sim\hat{P}_t$}. We now describe the procedure to obtain the dual variable $\bm{\lambda}_t$ for each period $t$.

We adopt the framework of expert problem, where we regard each constraint $i$ as an expert $i$ and we use the algorithm for the expert problem to update the dual variable. We regard $\bm{\lambda}_t$ as a distribution over the long-term constraints, after divided by a scaling factor $\mu$. Then, the range of the dual variable for all the minimax problems \eqref{eqn:121903} can be restricted to the set $\mu\cdot \Delta_m$ where $\Delta_m=\{\bm{y}\in\mathbb{R}_{\geq0}^m:\sum_{i=1}^{m}y_i=1 \}$ denotes a distribution over the long-term constraints and $\mu>0$ is a constant to be specified later. The player of the expert problem actually chooses one constraint $i_t$ among the long-term constraints, by setting $\bm{\lambda}_t=\mu\cdot\bm{e}_{i_t}$ where $\bm{e}_{i_t}\in\mathbb{R}^m$ is a vector with 1 as the $i_t$-th component and 0 for all other components. Clearly, what we can observe is a \textit{stochastic} outcome. We can observe the stochastic outcome $\hLInner(\bm{c}_t,\mu\cdot\bm{e}_{i_t},\theta_t)$ for the currently chosen constraint $i_t$. Finally, the second-stage decision $\bm{x}_t$ is determined by solving the inner minimization problem in \eqref{eqn:121903} for $\bm{c}=\bm{c}_t$ and $\bm{\lambda}=\mu\cdot\bm{e}_{i_t}$. Here, $\bm{c}_t$ is determined as the minimizer of $\hLInner_t(\bm{c}_t, \hat{\bm{\lambda}}^*)$ and $i_t$ is determined by the algorithm for the expert problem over the long-term constraints.

We now further specify what is the observed outcome for the dual player. Note that the dual player chooses a constraint $i_t$ as action. The corresponding outcome is $\hLInner_{i_t}(\bm{c}_t, \mu\cdot\bm{e}_{i_t},\theta_t)$, where $\hLInner_{i}(\bm{c}_t, \mu\cdot\bm{e}_{i_t},\theta_t)$ is defined as follows for each $i\in[m]$,
\begin{equation}\label{eqn:defOutISPnew}
\hLInner_{i}(\bm{c}_t, \mu\cdot\bm{e}_{i_t},\theta_t)=p(\bm{c}_t)-\frac{\mu}{T}+f_{\theta_t}(\bm{x}_t)+\frac{\mu\cdot g_{i,\theta_t}(\bc_t, \bm{x}_t)}{T\cdot\beta_i},
\end{equation}
where $\bm{x}_t$ denotes the second-stage decision we made at period $t$.
Though the action for the dual player is $\bm{e}_{i_t}$, we are actually able to obtain \textit{additional information} for the dual player, which helps the convergence of the expert algorithm. It is easy to see that we have the value of $\bLInner_i(\bm{c}_t,\mu\cdot\bm{e}_{i_t},\theta_t)$ at the end of period $t$, for all other constraint $i\neq i_t$. Therefore, for the expert algorithm, we are having the \textit{full feedback}, where the outcomes of all constraints, $\hLInner_i$ for all $i\in[m]$, can be observed. The above discussion implies that we can apply adversarially learning algorithm such as Hedge algorithm to dynamically update the dual variable $\bm{\lambda}_t$ for each period $t$. Our formal algorithm is described in \Cref{alg:IAL}, which we call \textit{Informative Adversarial Learning} (IAL) algorithm since the updates are informed by the predictions.

\begin{algorithm}[tb]
\caption{Informative Adversarial Learning (IAL) algorithm}
\label{alg:IAL}
\begin{algorithmic}
\STATE {\bfseries Input:} the scaling factor $\mu>0$, the adversarial learning algorithm $\ALGD$ for dual variable $\bm{\lambda}$, and the prior estimates $\hat{\bm{P}}$.
\STATE {\bfseries Initialize:} compute $\hat{\beta}_{i,t}$ for all $i\in[m], t\in[T]$ as \eqref{eqn:121902}.
\FOR{$t=1,\dots,T$}
\STATE $\bm{1}.$ $\ALGD$ returns a long-term constraint $i_t\in[m]$.
\STATE $\bm{2}.$ Set $\bm{c}_t=\text{argmin}_{\bm{c}\in\mathcal{C}}\hLInner_t(\bm{c}, \mu\cdot\bm{e}_{i_t})$.
\STATE $\bm{3}.$ Observe $\theta_t$ and set $\bm{x}_t$ by solving the inner problem in \eqref{eqn:121903} with $\bm{c}=\bm{c}_t$ and $\bm{\lambda}=\mu\cdot\bm{e}_{i_t}$.
\STATE $\bm{4}.$ Return to $\ALGD$ $\hLInner_{i,t}(\bm{c}_t, \mu\cdot\bm{e}_{i_t}, \theta_t)$ defined as follows for each $i\in[m]$,
\begin{align}
\hLInner_{i,t}(\bm{c}_t, \mu\cdot\bm{e}_{i_t}, \theta_t)=&p(\bm{c}_t)-\frac{\mu\cdot\hat{\beta}_{i,t}}{T\cdot\beta_i}+f_{\theta_t}(\bm{x}_t) \nonumber\\
&+\frac{\mu\cdot g_{i,\theta_t}(\bc_t, \bm{x}_t)}{T\cdot\beta_i}. \label{eqn:121916}
\end{align}
\ENDFOR
\end{algorithmic}
\end{algorithm}

As shown in the next theorem, the regret of \Cref{alg:IAL} is bounded by $O(W_T+\sqrt{T})$, which matches the lower bound established in \Cref{thm:priorLower}.

\begin{theorem}\label{thm:NStaRegretISP}
Denote by $\pi$ \Cref{alg:IAL} with input $\mu=\|\hat{\bm{\lambda}}^*\|_{\infty}=\alpha\cdot T$ for some constant $\alpha>0$. Denote by $W_T=\sum_{t=1}^{T}W(\hat{P}_t, P_t)$ the total measure of inaccuracy, with $W(\hat{P}_t, P_t)$ defined in \eqref{eqn:Wasserdist}. Then, under \Cref{assump:main}, the regret enjoys the upper bound
\begin{align}
\text{Regret}(\pi, T)=&\mathbb{E}_{\bm{\theta}\sim\bm{P}}[\ALG(\pi,\bm{\theta})]-\mathbb{E}_{\bm{\theta}\sim\bm{P}}[\ALG(\pi^*,\bm{\theta})] \nonumber\\
\leq& \tilde{O}(\sqrt{T\cdot\log m})+O( W_T )  \label{eqn:122001}
\end{align}
if Hedge is selected as $\ALGD$. Moreover, we have
\begin{align*}
\frac{1}{T}\sum_{t=1}^{T}g_{i,\theta_t}(\bc_t, \bm{x}_t)-\beta_i\leq \tilde{O}\left(\sqrt{\frac{\log m}{T}}\right)+O\left(\frac{W_T}{ T}\right), \label{eqn:122002}
\end{align*}
for each $i\in[m]$.
\end{theorem}
Notably, the inaccuracy of the prior estimates would result in a constraint violation that scales as $O(\frac{1}{\sqrt{T}}+\frac{W_T}{T})$ as shown in \Cref{thm:NStaRegretISP}. Therefore, as long as $W_T$ scales sublinearly in $T$, which guarantees a sublinear regret bound following \eqref{eqn:122001}, the constraint violation of our \Cref{alg:IAL} also scales as $o(1)$, implying that the solutions generated by \Cref{alg:IAL} is asymptotically feasible. Generating asymptotically feasible solutions is standard in OCOwC literature \citep{jenatton2016adaptive, neely2017online, yuan2018online,yi2021regret}.

\section{Extension to the Setting without Predictions}\label{sec:stationary}
In this section, we explore the setting where there are no machine-learned predictions and the true distributions are still unknown. If the true distributions can be arbitrarily non-stationary, then the setting becomes identical to the adversarial setting and in general, one cannot achieve a sublinear regret. Therefore, we consider a setting lying between the stationary and the non-stationary setting. To be specific, we assume that the true distributions are identical, i.e., $P_t=P$ for each $t\in[T]$. However, at each period $t$, after the type $\theta_t$ is realized, there can be an adversary corrupting $\theta_t$ into $\theta_t^c$, and only the value of $\theta^c_t$ is revealed to us. The adversarial corruption to a stochastic model can arise from the non-stationarity of the underlying distributions $\bm{P}$ (e.g. \citet{jiang2020online, balseiro2023best}), or malicious attack and false information input to the system (\citet{lykouris2018stochastic, gupta2019better}).

We first characterize the difficulty of the problem. Denote by $W(\bm{\theta})$ the total number of adversarial corruptions on sequence $\bm{\theta}$, i.e.,
\begin{equation}\label{eqn:Advercorrup}
  W(\bm{\theta})=\sum_{t=1}^{T}\bI(\theta_t\neq \theta_t^c).
\end{equation}
We show that the optimal regret bound scales at least $\Omega(\mathbb{E}_{\bm{\theta}\sim\bm{P}}[W(\bm{\theta})])$. Note that in the definition of the regret, the performance of our algorithm is the total collected reward \textit{after} adversarial corrupted, and the benchmark is the optimal policy \textit{with} adversarial corruptions, denoted by $\pi^*$.
\begin{theorem}\label{thm:adverLower}
Let $W(\bm{\theta})$ be the total number of adversarial corruptions on sequence $\bm{\theta}$, as defined in \eqref{eqn:Advercorrup}. For any feasible online policy $\pi$, there always there exists distributions $\bm{P}$ and a way to corrupt $\bm{P}$ such that
\begin{align}
\Reg^c(\pi, T)&=\mathbb{E}_{\bm{\theta}\sim\bm{P}}[\ALG(\pi,\bm{\theta}^c)]-\mathbb{E}_{\bm{\theta}\sim\bm{P}}[\ALG(\pi^*,\bm{\theta}^c)] \nonumber\\
&\geq
\Omega\left(\mathbb{E}_{\bm{\theta}\sim\bm{P}}[W(\bm{\theta})]\right). \nonumber
\end{align}
\end{theorem}
Our lower bound in \Cref{thm:adverLower} is in correspondence to the lower bound established in \cite{lykouris2018stochastic} for stochastic multi-arm-bandits model.

We now derive our algorithm to achieve a regret upper bound that matches the lower bound in \Cref{thm:adverLower} in terms of the dependency on $W(\bm{\theta})$. Our algorithm is modified from the previous \Cref{alg:IAL}. However, the caveat is that since we do not have the predictions $\hat{P}_t$ for each $t\in[T]$, we cannot directly obtain the value of $\bm{c}_t$ by minimizing over $\hLInner_t(\bm{c}, \bm{\lambda})$ in \eqref{eqn:121903}. One can indeed use historical samples to obtain an estimate of $P_t$ and plug the estimate into the formulation of $\hLInner_t(\bm{c}, \bm{\lambda})$ to solve it. However, the existence of the adversarial corruptions will bring new challenges to tackle with. Instead, in what follows, we seek for another approach that uses another adversarial learning algorithm to update $\bm{c}_t$.

We note that the minimax Lagrangian dual problem \eqref{lp:DualInner} admits the following reformulation, given the type realization $\theta$.
\begin{equation}\label{eqn:LISPstationary}
\begin{aligned}
\bLInner(\bm{c}, \bm{\lambda}, \theta)=&p(\bm{c})-\frac{1}{T}\cdot\sum_{i=1}^{m}\lambda_i+\min_{\bm{x}\in\mathcal{K}(\theta,\bm{c})}\left\{f_{\theta}(\bm{x})\right.  \\
&\left.+\sum_{i=1}^{m}\frac{\lambda_i\cdot g_{i,\theta}(\bc, \bm{x})}{T\cdot\beta_i}\right\}.
\end{aligned}
\end{equation}
We have the following lemma showing the convexity of $\bLInner(\bm{c}, \bm{\lambda}, \theta)$ over $\bm{c}$.
\begin{lemma}\label{lem:ConvexL}
For any $\bm{\lambda}$ and any $\theta$, $\bLInner(\bm{c}, \bm{\lambda}, \theta)$ defined in \eqref{eqn:LISPstationary} is a convex function over $\bm{c}$, under \Cref{assump:main}.
\end{lemma}
This above results imply that we can apply methods from online convex optimization (OCO) to update $\bm{c}_t$ for each $t\in[T]$.
The formal algorithm is presented in \Cref{alg:DAL}. Our algorithm admits a double adversarial learning structure. On the one hand, we use \textit{Online Gradient Descent} (OGD) algorithm \citep{zinkevich2003online} as $\ALG_1$ to update the first-stage decision $\bm{c}_t$. On the other hand, similar to \Cref{alg:IAL}, we regard each long-term constraint as an expert and use the expert algorithm such as \textit{Hedge} algorithm \citep{freund1997decision} as $\ALG_2$ to update the dual variable $\bm{\lambda}_t$, for each $t\in[T]$. Finally, the second-stage decision $\bm{x}_t$ is determined by solving the inner minimization problem in \eqref{eqn:LISPstationary} for $\bm{c}=\bm{c}_t$, $\bm{\lambda}=\bm{\lambda}_t$ and $\theta=\theta^c_t$.

\begin{algorithm}[tb]
\caption{Doubly Adversarial Learning (DAL) algorithm}
\label{alg:DAL}
\begin{algorithmic}
\STATE {\bfseries Input:} the scaling factor $\mu>0$, the adversarial learning algorithm $\ALG_1$ for the first-stage decision, the adversarial learning algorithm $\ALG_2$ for the dual variable.
\FOR{$t=1,\dots,T$}
\STATE $\bm{1}.$ $\ALG_1$ returns $\bm{c}_t$ and $\ALG_2$ returns $i_t\in[m]$.
\STATE $\bm{2}.$ Observe $\theta^c_t$ and determine $\bm{x}_t$ by solving the inner problem in \eqref{eqn:LISPstationary} with $\bm{c}=\bm{c}_t$ and $\bm{\lambda}=\mu\cdot\bm{e}_{i_t}$.
\STATE $\bm{3}.$ Return $\bLInner(\bm{c}_t,\mu\cdot\bm{e}_{i_t},\theta^c_t)$ to $\ALG_1$.
\STATE $\bm{4}.$ Return $\hLInner_i(\bm{c}_t,\mu\cdot\bm{e}_{i_t},\theta^c_t)$ defined in \eqref{eqn:defOutISPnew} for all $i\in[m]$ to $\ALG_2$.
\IF{$\frac{1}{T}\cdot\sum_{\tau=1}^{t}g_{i,\theta_{\tau}}(\bc_{\tau}, \bm{x}_{\tau})>\beta_i$ for some $i\in[m]$}
\STATE we terminate the algorithm by taking the null action $\bm{0}$ for both stage decision in the remaining horizon.
\ENDIF
\ENDFOR
\end{algorithmic}
\end{algorithm}

We now show that our \Cref{alg:DAL} achieves a regret bound that matches the lower bound in \Cref{thm:adverLower} in terms of the dependency on $W(\bm{\theta})$, which is in correspondence to the linear dependency on $W(\bm{\theta})$ established in \cite{gupta2019better} for the stochastic multi-arm-bandits model. 

\begin{theorem}\label{thm:CorruptRegretISP}
Denote by $\pi$ \Cref{alg:DAL} with input $\mu=T$. Then, under \Cref{assump:main} and the corrupted setting, the regret enjoys the upper bound
\begin{align}
\text{Regret}^{\text{c}}(\pi, T)&=\mathbb{E}_{\bm{\theta}\sim\bm{P}}[\ALG(\pi,\bm{\theta}^c)]-\mathbb{E}_{\bm{\theta}\sim\bm{P}}[\ALG(\pi^*,\bm{\theta}^c)] \nonumber\\
&\leq \tilde{O}((G+F)\cdot\sqrt{T})+\tilde{O}(\sqrt{T\cdot\log m}) \nonumber\\
&~~+O( \mathbb{E}_{\bm{\theta}\sim\bm{P}}[W(\bm{\theta})] ) , \label{eqn:121801}
\end{align}
if OGD is selected as $\ALG_1$ and Hedge is selected as $\ALG_2$. Here, $\mathbb{E}_{\bm{\theta}\sim\bm{P}}[W(\bm{\theta})]$ denotes the expectation of the total number of corruptions, with $W(\bm{\theta})$ defined in \eqref{eqn:Advercorrup}.
\end{theorem}

We remark that the implementation of our \Cref{alg:DAL} in this setting is \textit{agnostic} to the total number of adversarial corruptions and achieves the optimal dependency on the number of corruptions. This is one fascinating benefits of adopting adversarial learning algorithms as algorithmic subroutines in \Cref{alg:DAL}, where the corruptions can also be incorporated as adversarial input which is handled by the learning algorithms. On the other hand, even if the number of corruptions is given to us as a prior knowledge, \Cref{thm:adverLower} shows that still, no online policy can achieve a better dependency on the number of corruptions.

\section{Numerical Experiments}\label{sec:numerical}

In this section, we conduct numerical experiments to test the performance of our algorithms empirically.
We try two sets of experiments for the resource allocation problems.
One deals with the resource capacity constraints (packing constraints from a mathematical optimization view), where the long-term constraints take the formulation of $\sum_{t=1}^{T}\bm{g}(\bx_t, \bm{\theta}_t)\leq T\cdot\bm{\beta}$. The other set of experiments deals with the service level constraints (covering constraints from a mathematical optimization view), where the long-term constraints take the formulation of $\sum_{t=1}^{T}\bm{g}(\bx_t, \bm{\theta}_t)\geq T\cdot\bm{\beta}$. Note that the service level constraints (or the fairness constraints \citep{kearns2018preventing}) are widely studied in the literature (e.g. \cite{hou2009theory}), and we develop our algorithms to handle this type of covering constraints. The algorithmic development for the second set is described in \Cref{sec:Covering} in the supplementary material. In our experiments, we set the functions $\bm{g}(\bx_t, \bm{\theta}_t)$ to be linear functions over $\bx_t$.

\noindent\textbf{Experiment 1.} The first set deals with packing constraints. The offline problem can be formulated as follows:
\begin{align*}
&\max~~\sum_{t=1}^{T} c_t\\
&~~\mbox{s.t.}~~~\sum_{t=1}^{T}x_{i,t}\leq T\cdot \beta_i, ~~\forall i=1,\dots, 4\\
&~~~~~~~~~~x_{i,t}\leq D_{i,t,},~~\forall i, \forall t\\
&~~~~~~~~~~\sum_{i=1}^{4}x_{i,t}\geq \min\{ c_t, \sum_{i=1}^{4}D_{i,t} \},~~\forall t\\
&~~~~~~~~~~x_{i,t}\geq0, c_t\leq C,~~\forall i, \forall t.
\end{align*}
We do the numerical experiment under the following settings. Consider a resource allocation problem where there is 4 resources to be allocated. At each period, the decision maker needs to first decide a budget $c_t$ which restricts the total amount of units to be invested to each resources, and then, after the demand is realized, the decision maker needs to allocated the budget to satisfy the demand for each resource. The demand for each resource is normally distributed at each period (truncated at 0). We set the mean parameter $\mu_0=5$ and the standard deviation parameter $\sigma_0=10/3$. We set the vector $\bm{\beta}=[0.95, 0.90, 0.85, 0.80]$. We assume DAL is blind to the distributions while IAL knows the distributions. We test the performance of the IAL algorithm and DAL algorithm in the following four cases:
\begin{itemize}
  \item a. Stationary distribution case: demands of the four resources follow the same distribution $\mathcal{N}(k_0\mu_0, \sigma_0)$ with $k_0=2$.
  \item b. Non-stationary distribution case 1: We divide the whole horizon into two intervals: $[1, T/2]$ and $[T/2+1, T]$. During each interval, we sample the demands independently following the same distribution $\mathcal{N}(k_1\mu_0, \sigma_0)$ with $k_1=1, 3$ for the two intervals.
  \item c. Non-stationary distribution case 1: We divide the whole horizon into two intervals: $[1, T/2]$ and $[T/2+1, T]$. During each interval, we sample the demands independently following the same distribution $\mathcal{N}(k_2\mu_0, \sigma_0)$ with $k_2=3, 1$ for the two intervals.
  \item d. Non-stationary distribution case 3: We divide the whole horizon into five intervals: $[1, T/5]$, $[T/5+1, 2T/5]$, $[2T/5+1, 3T/5]$, $[3T/5+1, 4T/5]$, $[4T/5+1, T]$. During each interval, we sample the demands independently following the same distribution $\mathcal{N}(k_3\mu_0, \sigma_0)$ with $k_3=1, 2, 3, 2, 1$ for five intervals.
\end{itemize}
\textbf{Results and interpretations.} Note that the extent of non-stationarity is the same for the non-stationary case 1 and case 2. The difference is that the distribution of the two intervals is exchanged in case 1 and case 2. The extent of non-stationarity is largest for the non-stationary case 3. The numerical results are presented in \Cref{tab:01}. On one hand, for the stationary case, the performances of both DAL and IAL are good, with a relative regret within $4\%$. On the other hand, as we can see, the performance of DAL decays from situation (a) up to situation (d), as the non-stationarity of the underlying distributions gets larger and larger. However, the performance of IAL remains relatively stable because the non-stationarity of the underlying distributions has been well incorporated in the algorithmic design, which illustrates the effectiveness of IAL in a non-stationary environment.
\begin{table}[!h]
  \centering
  \begin{tabular}{|c|c|c|}
    \hline
     & stationary case & non-stationary case 1 \\
     \hline
    DAL & $3.43\%$ & $8.26\%$ \\
    \hline
    IAL & $3.45\%$  & $5.84\%$ \\
    \hline
    & non-stationary case 2 & non-stationary case 3 \\
    \hline
    DAL &  $8.26\%$ & $11.02\%$ \\
    \hline
    IAL & $5.86\%$ & $6.54\%$ \\
    \hline
  \end{tabular}
  \caption{The relative regret of DAL and IAL algorithms with capacity (packing) constraints}
  \label{tab:01}
\end{table}

\noindent\textbf{Experiment 2.} The second set considers the service level constraints, namely, the covering constraints. We consider the four cases which are exactly the same as the cases described in experiment 1. The other parameters are also set as the same as those in experiment 1, except that the long-term constraints now become $\sum_{t=1}^{T}\bm{g}(\bx_t, \bm{\theta}_t)\geq T\cdot\bm{\beta}$.

\begin{figure}[!h]
\centering
    \includegraphics[width=0.7\columnwidth]{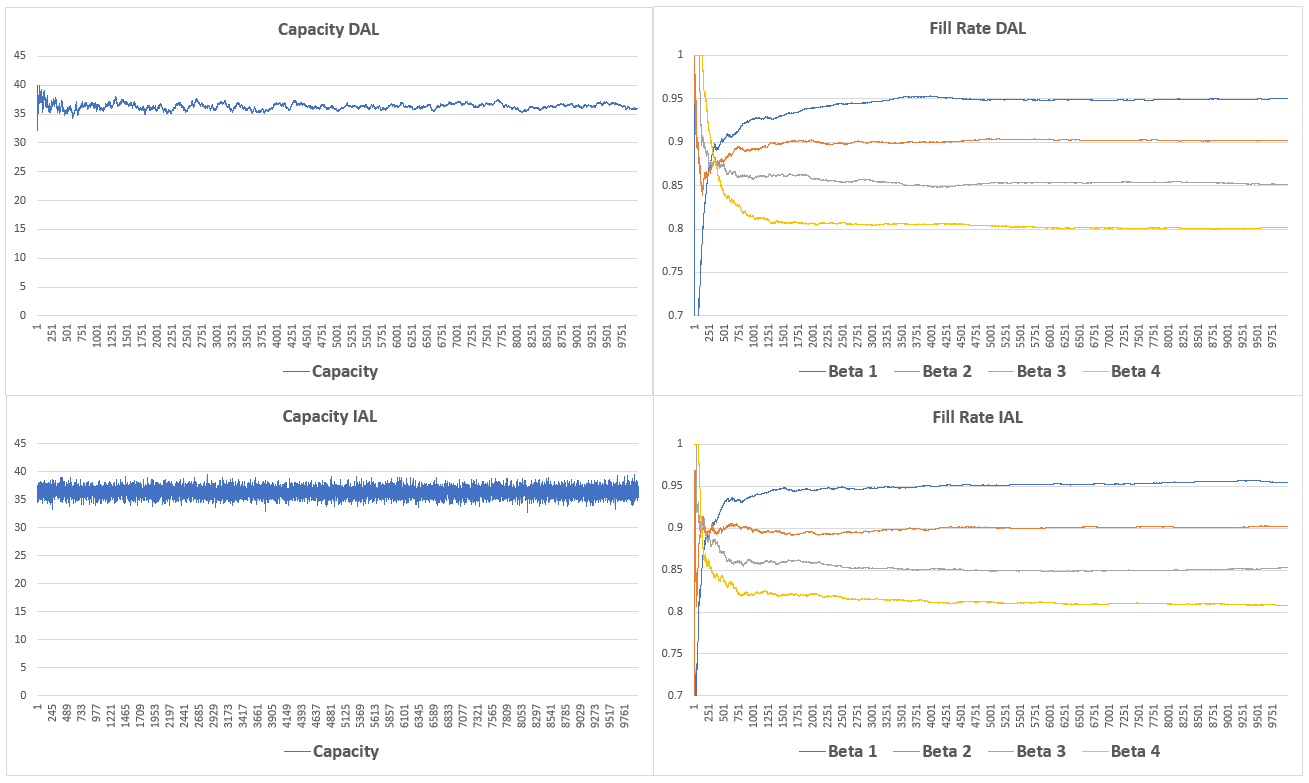}
    \caption{Numerical results of DAL and IAL algorithms with service level (covering) constraints for the stationary case.}
    \label{fig:a}
\end{figure}

\begin{figure}[!h]
\centering
    \includegraphics[width=0.7\columnwidth]{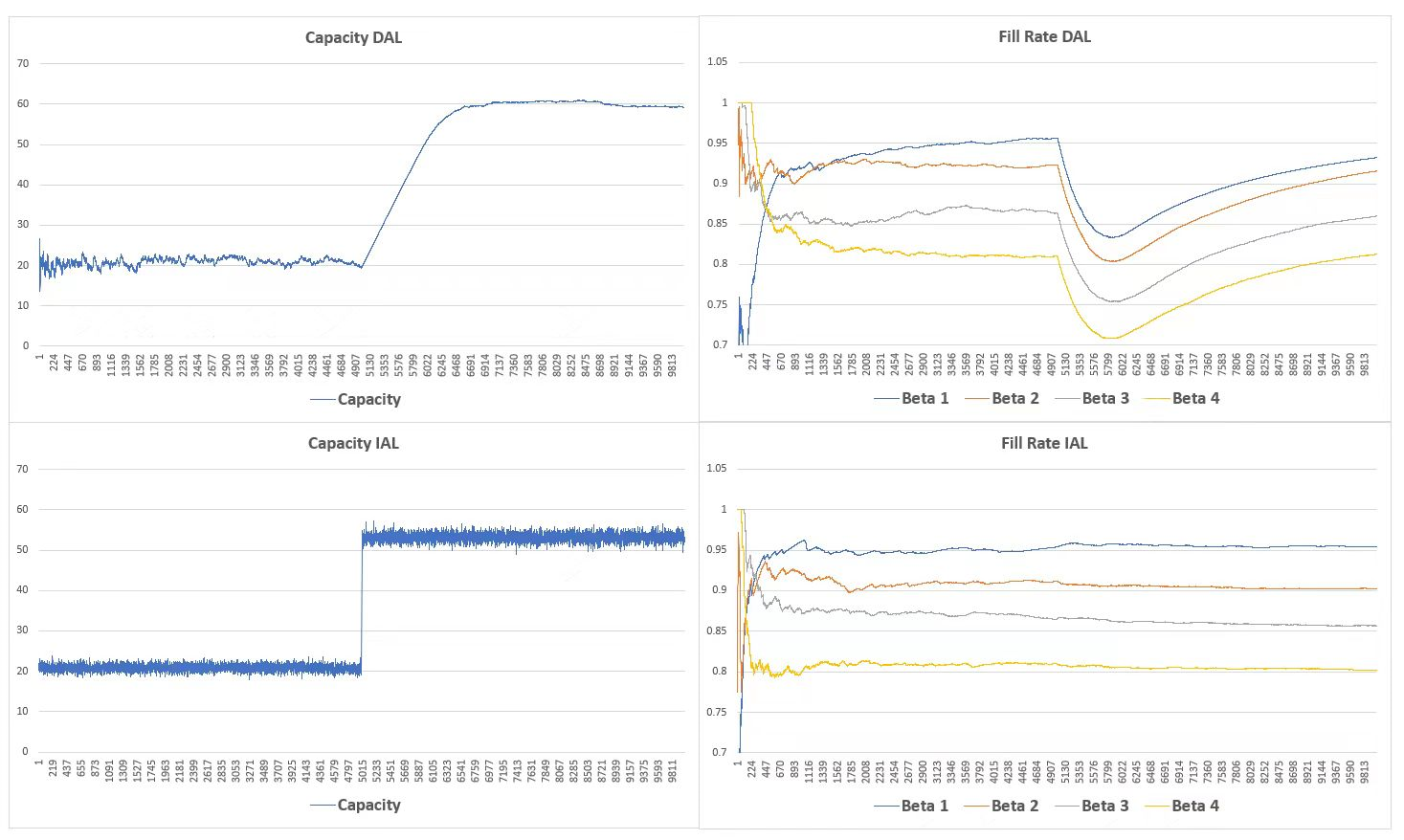}
    \caption{Numerical results of DAL and IAL algorithms with service level (covering) constraints for the non-stationary case 1.}
    \label{fig:b}
\end{figure}

\begin{figure}[!h]
\centering
    \includegraphics[width=0.7\columnwidth]{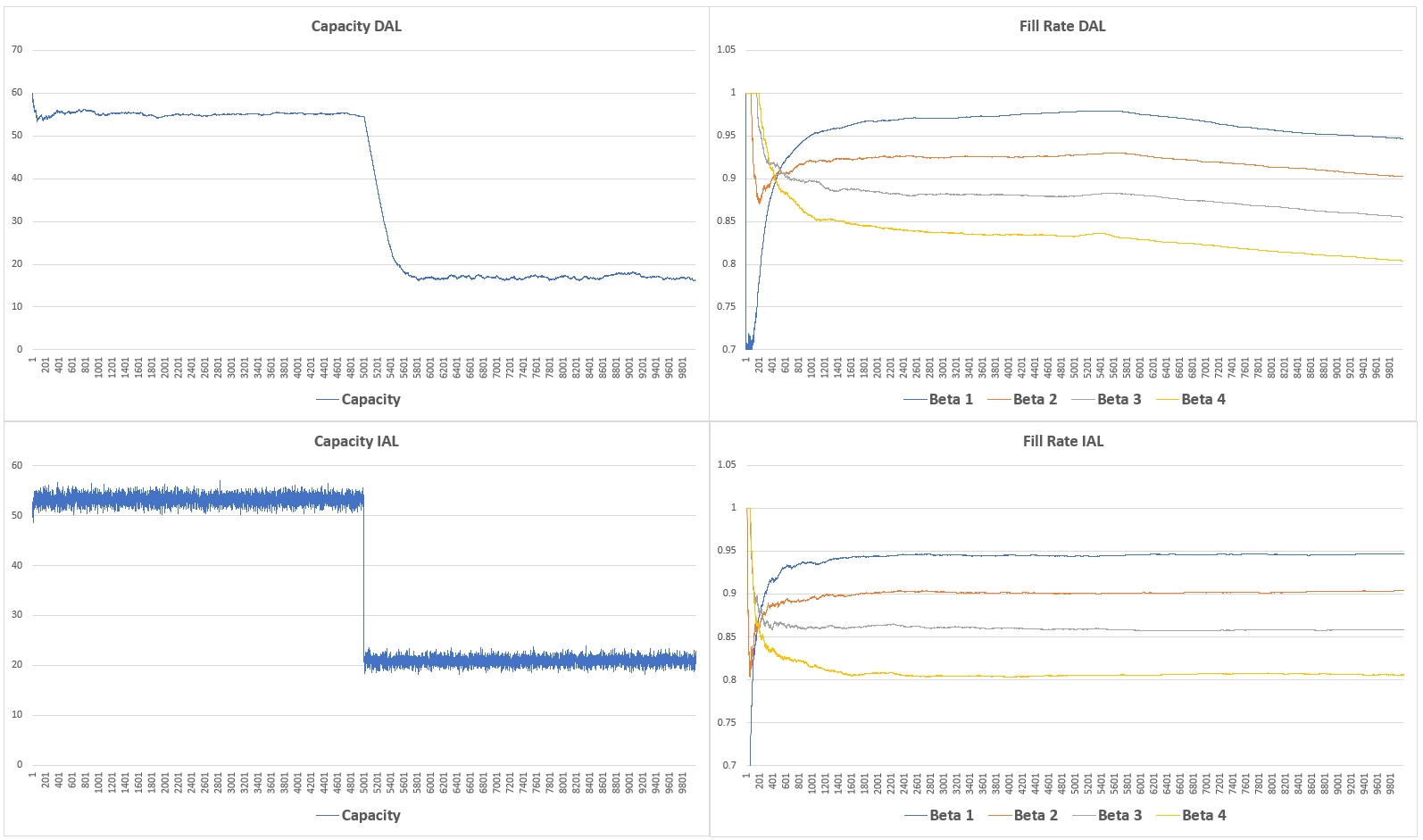}
    \caption{Numerical results of DAL and IAL algorithms with service level (covering) constraints for the non-stationary case 2.}
    \label{fig:c}
\end{figure}

\begin{figure}[!h]
\centering
    \includegraphics[width=0.7\columnwidth]{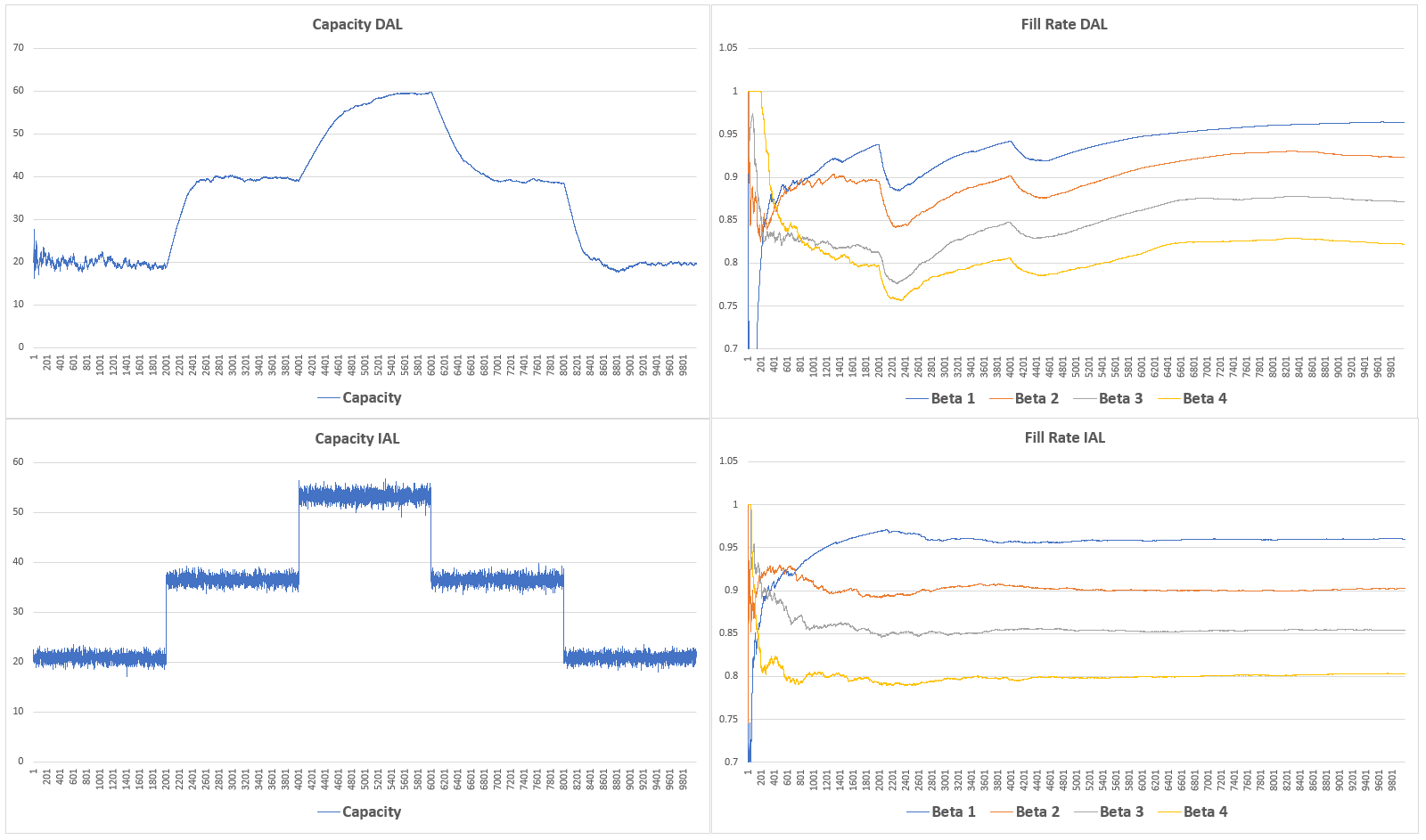}
    \caption{Numerical results of DAL and IAL algorithms with service level (covering) constraints for the non-stationary case 3.}
    \label{fig:d}
\end{figure}

\noindent\textbf{Results and interpretations.} In each of the figures, corresponding to each case, we plot 4 graphs capturing how the budget $c_t$ is determined and how the service level constraints are satisfied for each DAL and IAL during the horizon of $T=10000$ period. As we can see, both IAL and DAL converge rather quickly under the stationary case, in that the budget $c_t$ remains stable and the target service levels are achieved after about $1000$ periods, as plotted in \Cref{fig:a}. However, as we add non-stationarity in \Cref{fig:b}, \Cref{fig:c}, and \Cref{fig:d}, the performance of DAL decays in that the budget $c_t$ cannot capture the non-stationarity, and the achieved service levels are not stable. In contrast, for IAL, the budget $c_t$ changes quite quickly as long as the underlying distribution shifts. Moreover, even though the distributions changed during the horizon, the achieved service levels of IAL remain stable and the targets are reached. All these numerical results correspond to our theoretical finding and illustrate the benefits of IAL under non-stationarity.

\centering

\raggedright

\section{Summary}\label{sec:summary}

This paper proposes and studies the problem of bounding regret for online two-stage stochastic optimization with long-term constraints. The main contribution is an algorithmic framework that develops new algorithms via adversarial learning algorithms. The framework is applied to various setting. For the stationary setting, the resulted DAL algorithm is shown to achieve a sublinear regret, with a performance robust to adversarial corruptions. For the non-stationary (adversarial) setting, a modified IAL algorithm is developed, with the help of prior estimates. The sublinear regret can also be acheived by IAL algorithm as long as the cumulative inaccuracy of the prior estimates is sublinear.

\bibliographystyle{abbrvnat}
\bibliography{bibliography}

\clearpage

%
%
%

\begin{APPENDICES}
\crefalias{section}{appendix}

\section{Further Related Literature}\label{sec:RelatedWork}
Our model synthesizes the bandits-based model, where the representative problems include MAB problem, bandits with knapsack (BwK) problem and the more general OCO problem, and the feature-based model, where the representative problems include online allocation problem and a special case online packing problem. We now review these literature.

One representative problem for the bandits-based model is the BwK problem which reduces to MAB problem if no long-term constraints. The previous BwK results have focused on a stochastic setting \citep{badanidiyuru2013bandits, agrawal2014bandits}, where a $O(\sqrt{T})$ regret bound has been derived, and an adversarial setting (e.g. \citet{rangi2018unifying, immorlica2019adversarial}), where the sublinear regret is impossible to obtain and a $O(\log T)$ competitive ratio has been derived. Similar to our DAL algorithm, the algorithms in the previous literature reply on an interplay between the primal and dual LPs. To be more concrete, \citet{agrawal2014bandits} and \citet{agrawal2014fast} develop a dual-based algorithms to BwK and a more general online stochastic optimization problem (belonging to bandits-based model) and analyzes the algorithm performance under further stronger conditions on the dual optimal solution. However, these learning models and algorithms are developed in the stationary (stochastic) environment, which cannot be applied to the non-stationary setting. For the non-stationary setting, the recent work \cite{liu2022non} derives sublinear regret, based on a more involved complexity measure over the non-stationarity of the underlying distributions which concerns both the temporary changes of two neighborhood distributions and the global changes of the entire distribution sequence.

Another representative problem for the bandits-based model is the OCO problem, which is one of the leading online learning frameworks \citep{hazan2016introduction}. Note that the standard OCO problem generally adopts a static optimal policy as the benchmark, i.e., the decision of the benchmark needs to be the same for each period. In contrast, in our model, the benchmark is a more powerful dynamic optimal policy where the decisions are allowed to be non-homogeneous across time. Therefore, not only our model is more involved, our benchmark is also stronger.
There have been results that consider a dynamic optimal policy as the benchmark for OCO \citep{besbes2015non, hall2013dynamical, jadbabaie2015online}, but all these works consider the unconstrained setting with no long-term constraints. For the line of works that study the problem of online convex optimization with constraints (OCOwC), existing literature would assume the constraint functions that characterize the long-term constraints are either static \citep{jenatton2016adaptive,yuan2018online,yi2021regret} or stochastically generated \citep{neely2017online}.

For the type-based model, one representative problem is the online packing problem, where the columns and the corresponding coefficient in the objective of the underlying LP come one by one and the decision has to be made on-the-fly. The packing problem covers a wide range of applications, including secretary problem \citep{ferguson1989solved, arlotto2019uniformly}, online knapsack problem \citep{arlotto2020logarithmic, jiang2019online}, resource allocation problem \citep{li2020simple}, network routing problem \citep{buchbinder2009online}, matching problem \citep{mehta2007adwords} etc. The problem is usually studied under either a stochastic model where the reward and size of each query is drawn independently from an unknown distribution $\mathcal{P}$, or a more general the random permutation model where the queries arrive in a random order  \citep{molinaro2014geometry, agrawal2014dynamic, kesselheim2014primal, gupta2014experts}. The more general online allocation problem (e.g. \cite{balseiro2022best}) has also been considered in the literature, where the objective and the constraint functions are allowed to be general functions.

\section{Extensions}\label{sec:extensions}

\subsection{Non-convex Objective and Non-concave Constraints}
Note that in \Cref{assump:main}, we require a convexity condition where the function $p(\cdot)$ is a convex function with $p(\bm{0})=0$ and $\mathcal{C}$ is a compact and convex set which contains $\bm{0}$. Also, the functions $f_{\theta}(\cdot)$ need to be convex in $\bm{x}$ and $\bm{g}_{\theta}(\cdot, \cdot)$ need to be convex in both $\bc$ and $\bm{x}$, for any $\theta$. In this section, we explore whether these convexity conditions can be removed with further characterization of our model.

We now show that when $\mathcal{C}$ contains only a finite number of elements, we can remove all the convexity requirements in \Cref{assump:main} and we still obtain a $\tilde{O}(\sqrt{T}+W_T)$ regret bound for the setting in \Cref{sec:stationary}. As for the setting in \Cref{sec:nonstationary}, we can obtain the same $\tilde{O}(\sqrt{T}+W_T)$ regret bound for \textit{arbitrary} $\mathcal{C}$, as long as the minimization step $\bm{c}_t=\text{argmin}_{\bm{c}\in\mathcal{C}}\hLInner_t(\bm{c}, \mu\cdot\bm{e}_{i_t})$ can be efficiently solved in \Cref{alg:IAL}.

We now assume $\mathcal{C}$ has a finite support and remove all the convexity requirements in \Cref{assump:main}. In order to recover the results in \Cref{sec:stationary}, we regard each element in $\mathcal{C}$ as an expert and we apply expert algorithms to decide $\bc_t$ for each period $t$. To be more concrete, note that for the first-stage decision, we only have \textit{bandit-feedback}, i.e., we can only observe the stochastic outcome of selecting $\bc_t$ instead of other elements in the set $\mathcal{C}$. Therefore, we can apply EXP3 algorithm \citep{auer2002nonstochastic} as $\ALG_1$. Still, following the same procedure as the proof of \Cref{thm:CorruptRegretISP}, we can reduce the regret bound of \Cref{alg:DAL} into the regret bound of $\ALG_1$, $\ALG_2$, and an additional $O(W_T)$ term if adversarial corruption exists. The above argument is formalized in the following theorem.
\begin{theorem}\label{thm:DiscreteC}
Denote by $\pi$ \Cref{alg:DAL} with input $\mu=T$. Then, if $\mathcal{C}$ constains only a finite number of elements, denoted by $K$, with condition c in \Cref{assump:main}, it holds that
\begin{align}
\text{Regret}^{\text{c}}(\pi, T)=\mathbb{E}_{\bm{\theta}\sim\bm{P}}[\ALG(\pi,\bm{\theta}^c)]-\mathbb{E}_{\bm{\theta}\sim\bm{P}}[\ALG(\pi^*,\bm{\theta}^c)] \leq &\tilde{O}(\sqrt{K\cdot T})+\tilde{O}(\sqrt{T\cdot\log m})\nonumber\\
&+O( \mathbb{E}_{\bm{\theta}\sim\bm{P}}[W(\bm{\theta})] ) , \label{eqn:012501}
\end{align}
if EXP3 is selected as $\ALG_1$ and Hedge is selected as $\ALG_2$. Here, $\mathbb{E}_{\bm{\theta}\sim\bm{P}}[W(\bm{\theta})]$ denotes the expectation of the total number of corruptions, with $W(\bm{\theta})$ defined in \eqref{eqn:Advercorrup}.
\end{theorem}

We now consider the setting in \Cref{sec:nonstationary} and we show that the $\tilde{O}(\sqrt{T}+W_T)$ regret bound holds for \Cref{alg:IAL} without the convexity requirements in \Cref{assump:main}, for arbitrary $\mathcal{C}$ as long as the minimization step $\bm{c}_t=\text{argmin}_{\bm{c}\in\mathcal{C}}\hLInner_t(\bm{c}, \mu\cdot\bm{e}_{i_t})$ can be efficiently solved. The key idea is to regard the first-stage decision at each period as a distribution over $\mathcal{C}$, denoted by $\tbc_t$. Then, $\mathbb{E}_{\tilde{\bm{C}}}\left[\hat{L}(\tilde{\bm{C}}, \bm{\lambda})\right]$ would be a convex function over the distribution $\tilde{\bm{C}}$. Moreover, for each fixed $\bm{\lambda}$, note that selecting the worst distribution $\tilde{\bm{C}}$ to minimize $\mathbb{E}_{\tilde{\bm{C}}}\left[\hat{L}(\tilde{\bm{C}}, \bm{\lambda})\right]$ can be reduced equivalent to selecting the worst deterministic $\bm{C}$ to minimize $\hat{L}(\bm{C}, \bm{\lambda})$. Therefore, every step of the proof of \Cref{lem:decompose} and \Cref{thm:NStaRegretISP} would follow by replacing $\bc_t$ into the distribution $\tbc_t$ for each $t\in[T]$. We have the following result.
\begin{theorem}\label{thm:NconvexNStaRegretISP}
Denote by $\pi$ \Cref{alg:IAL} with input $\mu=\|\hat{\bm{\lambda}}^*\|_{\infty}=\alpha\cdot T$ for some constant $\alpha>0$. Denote by $W_T=\sum_{t=1}^{T}W(\hat{P}_t, P_t)$ the total measure of inaccuracy, with $W(\hat{P}_t, P_t)$ defined in \eqref{eqn:Wasserdist}. Then, under condition c of \Cref{assump:main}, for arbitrary $\mathcal{C}$, the regret enjoys the upper bound
\begin{align}
\text{Regret}(\pi, T)=&\mathbb{E}_{\bm{\theta}\sim\bm{P}}[\ALG(\pi,\bm{\theta})]-\mathbb{E}_{\bm{\theta}\sim\bm{P}}[\ALG(\pi^*,\bm{\theta})]
\leq \tilde{O}(\sqrt{T\cdot\log m})+O( W_T )  \label{eqn:012509}
\end{align}
if Hedge is selected as $\ALGD$. Moreover, we have
\begin{align}
\frac{1}{T}\sum_{t=1}^{T}g_{i,\theta_t}(\bc_t, \bm{x}_t)-\beta_i\leq \tilde{O}\left(\sqrt{\frac{\log m}{T}}\right)+O\left(\frac{W_T}{ T}\right), \label{eqn:012510}
\end{align}
for each $i\in[m]$.
\end{theorem}

\subsection{Incorporating Covering Constraints}\label{sec:Covering}

In this section, we discuss how to incorporate covering constraints into our model. Note that in the previous section, we let the long-term constraints $\frac{1}{T}\cdot\sum_{t=1}^{T}\bx_t\in\mathcal{B}(\bm{C}, \bm{\theta})$ be characterized as follows,
\begin{equation}\label{eqn:012511}
  \frac{1}{T}\cdot\sum_{t=1}^{T}\bm{g}_{\theta_t}(\bc_t, \bm{x}_t)\leq\bm{\beta},
\end{equation}
with $\bm{\beta}\in(0,1)^m$, which corresponds to packing constraints since both $\bm{g}(\cdot)$ and $\bm{\beta}$ are non-negative. For the case with covering constraints, the long-term constraints $\frac{1}{T}\cdot\sum_{t=1}^{T}\bx_t\in\mathcal{B}(\bm{C}, \bm{\theta})$ would enjoy the following characterization
\begin{equation}\label{eqn:012512}
  \frac{1}{T}\cdot\sum_{t=1}^{T}\bm{g}_{\theta_t}(\bc_t, \bm{x}_t)\geq\bm{\beta},
\end{equation}
with $\bm{\beta}\in(0,1)^m$. We now show that all our previous results hold for the packing constraints \eqref{eqn:012512}, by simply changing the input parameter $\mu$ in \Cref{alg:DAL} and \Cref{alg:IAL}. We illustrate through the stationary setting consiered in \Cref{sec:stationary} where $P_t=P$ for each $t\in[T]$.

Now, \eqref{lp:OPT} enjoys the following new formulation.
\begin{align}
 \OPTOuter= \min & \sum_{t=1}^{T}\mathbb{E}_{\tbc_t, \tbx_t, \theta_t}\left[p(\tbc_t)+f_{\theta_t}(\tbx_t)\right] \label{lp:Outer}\\
  \mbox{s.t.} & \frac{1}{T}\cdot\sum_{t=1}^{T}\mathbb{E}_{\tbc_t, \tbx_t, \theta_t}[\bm{g}_{\theta_t}(\tbc_t, \tbx_t)]\geq\bm{\beta} \nonumber\\
  & \tbx_t\in\mathcal{K}(\theta_t, \tbc_t), \tbc_t\in\mathcal{C}, \forall t.\nonumber
\end{align}
The Lagrangian dual of $\OPTOuter$ \eqref{lp:Outer} can be formulated as follows.
\begin{equation}\label{lp:Dualouter}
  \max_{\bm{\lambda}\geq0}\min_{\tbc_t\in\mathcal{C}} \LOuter(\bm{C}, \bm{\lambda})=\sum_{t=1}^{T}\mathbb{E}\left[p(\tbc_t)+\frac{1}{T}\cdot\sum_{i=1}^{m}\lambda_i+\min_{\tbx_t\in\mathcal{K}(\theta_t,\tbc_t)}f_{\theta_t}(\tbx_t)-\sum_{i=1}^{m}\frac{\lambda_i\cdot g_{i,\theta_t}(\tbc_t, \tbx_t)}{T\cdot\beta_i}\right].
\end{equation}
Similar to \eqref{eqn:LISPstationary}, we have the following formulation for $\bLInner(\bm{c}, \bm{\lambda}, \theta)$ as the single-period decomposition of $ \LOuter(\bm{C}, \bm{\lambda})$.
\begin{equation}\label{eqn:LOSPstationary}
\bLOuter(\bm{c}, \bm{\lambda}, \theta)=p(\bm{c})+\frac{1}{T}\cdot\sum_{i=1}^{m}\lambda_i+\min_{\bm{x}\in\mathcal{K}(\theta,\bm{c})}\left\{f_{\theta}(\bm{x})-\sum_{i=1}^{m}\frac{\lambda_i\cdot g_{i,\theta}(\bc, \bm{x})}{T\cdot\beta_i}\right\}.
\end{equation}
We can obtain the following result.
\begin{lemma}\label{lem:NEWStaDual}
Under the stationary setting where $P_t=P$ for each $t\in[T]$, it holds that
\begin{equation*}
\max_{\bm{\lambda}\geq0}\min_{\tbc_t\in\mathcal{C}} \mathbb{E}_{\tilde{\bm{C}}}[\LOuter(\bm{C}, \bm{\lambda})]=\max_{\bm{\lambda}\geq0}\min_{\tbc\in\mathcal{C}} T\cdot\mathbb{E}_{\tbc, \theta\sim P}\left[\bLOuter(\tbc, \bm{\lambda}, \theta)\right].
\end{equation*}
\end{lemma}
Therefore, we can still regard solving the dual problem \eqref{lp:Dualouter} as a procedure of solving the repeated zero-sum games, where player 1 chooses the first-stage decision $\bc_t$ and player 2 chooses one long term constraint $i_t\in[m]$. We still apply \Cref{alg:DAL} with the new definition
\begin{equation}\label{eqn:defOutOSP}
\bLOuter_{i}(\bm{c}_t, \mu\cdot\bm{e}_{i_t},\theta_t)=p(\bm{c}_t)+\frac{\mu}{T}+f_{\theta_t}(\bm{x}_t)-\frac{\mu\cdot g_{i,\theta_t}(\bc_t, \bm{x}_t)}{T\cdot\beta_i}.
\end{equation}
However, in order to incorporate covering constraints, we need to adjust the value of the scaling factor $\mu$ and we only terminate our algorithm after the entire horizon has been run out. In fact, we need an upper bound (arbitrary upper bound suffices) over the $l_1$ norm of the optimal dual variable $\bm{\lambda}^*$ of \eqref{lp:Dualouter} to serve as the scaling factor $\mu$ and in practice, we can spend the first $\sqrt{T}$ time periods to construct an upper confidence interval of $\bm{\lambda}^*$ to serve as $\mu$ without influencing the order of the regret bound.

\begin{theorem}\label{thm:RegretOSP}
Denote by $\pi$ \Cref{alg:DAL} with input $\mu$ being an upper bound of $\|\bm{\lambda}^*\|_1$ where $\bm{\lambda}^*$ is the optimal dual variable of \eqref{lp:Dualouter}. Then, under \Cref{assump:main}, $\mu=\alpha\cdot T$ for some constant $\alpha>0$ and if OGD is selected as $\ALG_1$ and Hedge is selected as $\ALG_2$, the regret enjoys the upper bound
\begin{equation}\label{eqn:121501}
\text{Regret}^{\text{TSC}}(\pi, T)\leq \tilde{O}((G+F)\cdot\sqrt{T})+\tilde{O}(\sqrt{T\cdot\log m})
\end{equation}
and moreover, we have
\begin{equation}\label{eqn:121502}
\beta_i-\frac{1}{T}\cdot\sum_{t=1}^{T}g_{i,\theta_t}(\bc_t, \bm{x}_t)\leq \tilde{O}\left(\frac{G+F}{\alpha\sqrt{T}}\right)+\tilde{O}\left(\frac{\sqrt{\log m}}{\alpha\cdot\sqrt{T}}\right) ,~~\forall i\in[m].
\end{equation}
\end{theorem}

\section{Regret Bounds for Adversarial Learning}

In this section, we present the implementation details of two adversarial learning algorithms, OGD and Hedge, that will be used as algorithmic subroutines in our \Cref{alg:DAL} and \Cref{alg:IAL}, as well as their regret analysis.

OGD is an algorithm to be executed in a finite horizon of $T$ periods, and at each period $t$, OGD selects an action $\bm{c}_t\in\mathcal{C}$, receives an adversarial chosen cost function $h_t(\cdot)$ afterwards, and incurs a cost $h_t(\bm{c}_t)$. OGD is designed to minimize the regret
\[
\text{Reg}_{\text{OGD}}(T)=\sum_{t=1}^{T}h_t(\bm{c}_t)-\min_{\bm{c}\in\mathcal{C}}\sum_{t=1}^{T}h_t(\bm{c}).
\]
The implementation of OGD is described in \Cref{alg:OGD}. In our problem, $h_t=\bar{L}^{\text{ISP}}_i(\bm{c}_t,\mu\cdot\bm{e}_{i_t},\theta_t)$ for ISP and $h_t=\bar{L}^{\text{OSP}}_i(\bm{c}_t,\mu\cdot\bm{e}_{i_t},\theta_t)$ for OSP. In order to obtain a subgradient, one can compute the optimal dual variable of the inner minimization problem of \eqref{eqn:LISPstationary}. For example, the inner minimization problem is
\begin{align}
  \min &~~~ f_{\theta_t}(\bm{x})+\frac{\mu\cdot g_{i_t,\theta_t}(\bc, \bm{x})}{T\cdot\beta_{i_t}} \label{lp:InnerISP}\\
  \mbox{s.t.} &~~~ B_{\theta_t}\bm{x}\leq \bm{c}_t\nonumber\\
  &~~~ \bm{x}\geq0 \nonumber
\end{align}
and the Lagrangian dual problem of \eqref{lp:InnerISP} is
\[
\max_{\bm{\gamma}\leq0} \bm{\gamma}^\top\bm{c_t}+\min_{\bm{x}\geq0}f_{\theta_t}(\bm{x})-\bm{\gamma}^\top B_{\theta_t}\bm{x}+\frac{\mu\cdot g_{i_t,\theta_t}(\bc_t, \bm{x})}{T\cdot\beta_{i_t}}.
\]
The optimal dual solution $\bm{\gamma}_t^*$ can be computed from the above minimax problem and we know that
\[
\nabla h_t(\bm{c}_t)=\bm{\gamma}_t^*+\frac{\mu\cdot \nabla_{\bc_t} g_{i_t,\theta_t}(\bc_t, \bm{x})}{T\cdot\beta_{i_t}}.
\]
Clearly when $\mu=a\cdot T$ for a constant $a>0$, we have $\|\bm{\gamma}_t^*\|_2\leq G$ for some constant $G$ that depends only on (the upper bound of gradients of) $\{f_{\theta}, \bm{g}_{\theta}\}_{\forall \theta}$, the minimum positive element of $B_{\theta}$ for all $\theta$, and $\max_{i\in[m]}\{\frac{1}{\beta_i}\}$. The regret bound of OGD is as follows.
\begin{theorem}[Theorem 1 of \cite{zinkevich2003online}]\label{thm:OGDregret}
If $\eta_t=\frac{1}{\sqrt{t}}$, then it holds that
\[
\text{Reg}_{\text{OGD}}(T)\leq O\left((G+F)\cdot\sqrt{T}\right)
\]
where $F$ is an upper bound of the diameter of the set $\mathcal{C}$ and $G$ is a constant that depends only on (the upper bound of gradients of) $\{f_{\theta}, \bm{g}_{\theta}\}_{\forall \theta}$, the minimum positive element of $B_{\theta}$ for all $\theta$, and $\max_{i\in[m]}\{\frac{1}{\beta_i}\}$.
\end{theorem}

\begin{algorithm}[tb]
\caption{Online Gradient Descent (OGD) algorithm}
\label{alg:OGD}
\begin{algorithmic}
\STATE {\bfseries Input:} the step size $\eta_t$ for each $t\in[T]$.
\STATE Initially set an arbitrarily $\bm{c}_1\in\mathcal{C}$.
\FOR{$t=1,\dots,T$}
\STATE $\bm{1}.$ Take the action $\bm{c}_t$.
\STATE $\bm{2}.$ Observe the cost function $h_t(\cdot)$.
\STATE $\bm{3}.$ Update action
\[
\bm{c}_{t+1}=\mathcal{P}_{\mathcal{C}}\left(\bm{c}_t-\eta_t\cdot \nabla h_t(\bm{c}_t)\right)
\]
where $\nabla h_t(\bm{c}_t)$ denotes a subgradient of $h_t$ at $\bm{c}_t$ and $\mathcal{P}_{\mathcal{C}}$ denotes a projection to the set $\mathcal{C}$.
\ENDFOR
\end{algorithmic}
\end{algorithm}

The Hedge algorithm is used to solve the expert problem in a finite horizon of $T$ periods. There are $m$ experts and at each period $t$, Hedge will select one expert $i_t\in[m]$ ($i_t$ can be randomly chosen), observe the reward vector $\bm{l}_t\in\mathbb{R}^m$ afterwards, and obtain an reward $l_{i_t,t}$. Hedge is designed to minimize the regret
\[
\text{Reg}_{\text{Hedge}}(T)=\max_{i\in[m]}\sum_{t=1}^{T}l_{i,t}-\sum_{t=1}^{T}\mathbb{E}_{i_t}[l_{i_t, t}].
\]
The Hedge algorithm is described in \Cref{alg:Hedge}. In our problem, for ISP, we have
\[
l_{i,t}=\bar{L}_{i}(\bm{c}_t, \mu\cdot\bm{e}_{i_t},\theta_t).
\]
Under \Cref{assump:main}, when $\mu=\alpha\cdot T$ for a constant $\alpha$, we know that there exists a constant $\delta>0$ such that $|l_{i,t}|\leq\delta$, for all $i\in[m]$ and $t\in[T]$. Here, $\delta$ depends on $\max_{i\in[m]}\{\frac{1}{\beta_i}\}$.
\begin{theorem}[from Theorem 2 in \cite{freund1997decision}]\label{thm:Hedgeregret}
If $\eps=\sqrt{\frac{\log m}{T}}$, then it holds that
\[
\text{Reg}_{\text{Hedge}}(T)\leq \tilde{O}(\sqrt{T\cdot\log (m)})
\]
where the constant term in $\tilde{O}(\cdot)$ depends on $\max_{i\in[m]}\{\frac{1}{\beta_i}\}$.
\end{theorem}

\begin{algorithm}[tb]
\caption{Hedge algorithm}
\label{alg:Hedge}
\begin{algorithmic}
\STATE {\bfseries Input:} a parameter $\eps>0$.
\STATE {\bfseries Initialize:} $\bm{w}_1=\bm{1}\in\mathbb{R}^m$ and $\bm{y}_1=\frac{1}{m}\cdot\bm{w}_1$.
\FOR{$t=1,\dots,T$}
\STATE $\bm{1}.$ Take the action $i_t\sim\bm{y}_t$.
\STATE $\bm{2}.$ Observe the reward vector $\bm{l}_t$ and obtain a reward $l_{i_t, t}$.
\STATE $\bm{3}.$ Update the weight
\[
w_{i,t+1}=w_{i,t}\cdot\exp(-\eps\cdot l_{i,t})
\]
for each $i\in[m]$ and set
\[
y_{i,t+1}=\frac{w_{i,t+1}}{\sum_{i'=1}^{m}w_{i',t+1}}
\]
for each $i\in[m]$.
\ENDFOR
\end{algorithmic}
\end{algorithm}

\section{Missing Proofs for Section \ref{sec:nonstationary}}

\begin{myproof}[Proof of \Cref{thm:priorLower}]
The proof is modified from the proof of \Cref{thm:adverLower}.
We consider a special case of our problem where for any $\bc\in\mathcal{C}$ and any $\theta$, $\mathcal{K}(\theta, \bc)=[0,1]$ (there is no need to decide the first-stage decision). There is only one long-term constraint with target $\beta=\frac{1}{2}$. Moreover, there are three possible values of $\theta$, denoted by $\{\theta^1, \theta^2, \theta^3\}$. We have $f_{\theta^1}(x)=-x$, $f_{\theta^2}(x)=-\left(1+\frac{W_T}{T}\right)x$, $f_{\theta^3}(x)=-\left(1-\frac{W_T}{T}\right)x$ and $g_{\theta^1}(x)=g_{\theta^2}(x)=g_{\theta^3}(x)=x$ (only one long-term constraint). The prior estimate is $\hat{P}_t=\theta^1$ with probability 1 for each $t\in[T]$, and the problem with respect to the prior estimates can be described below in \eqref{appendixeg0}.
\begin{align}
   \min \ \ &  -x_1-...-x_{\frac{T}{2}}-x_{\frac{T}{2}+1}-...-x_{T}  \label{appendixeg0} \\
    \text{s.t. }\ & x_1+...+x_{\frac{T}{2}}+x_{\frac{T}{2}+1}+...+x_{T} \leq \frac{T}{2}\nonumber \\
    & 0 \leq x_t \leq 1\ \text{ for } t=1,...,T. \nonumber
\end{align}
Now we consider the following two possible true distributions. The first possible true scenario, given in \eqref{appendixeg3}, is that the distribution  $P_t=\theta^1$ for $t=1,\dots,\frac{T}{2}$ and $P_t=\theta^2$ for $t=\frac{T}{2}+1,\dots,T$. The second possible true scenario, given in \eqref{appendixeg4}, is that the distribution $P_t=\theta^1$ for $t=1,\dots,\frac{T}{2}$ and $P^c_t=\theta^3$ for $t=\frac{T}{2}+1,\dots,T$.
\begin{align}
   \min \ \ &  -x_1-...-x_{\frac{T}{2}}-\left(1+\frac{W_T}{T}\right)x_{\frac{T}{2}+1}-...-\left(1+\frac{W_T}{T}\right)x_{T}  \label{appendixeg3} \\
    \text{s.t. }\ & x_1+...+x_{\frac{T}{2}}+x_{\frac{T}{2}+1}+...+x_{T} \le \frac{T}{2}\nonumber \\
    & 0 \leq x_t \leq 1\ \text{ for } t=1,...,T. \nonumber \\
   \min \ \ &  -x_1-...-x_{\frac{T}{2}} -\left(1-\frac{W_T}{T}\right)x_{\frac{T}{2}+1}-...-\left(1-\frac{W_T}{T}\right)x_{T} \label{appendixeg4} \\
    \text{s.t. }\ & x_1+...+x_{\frac{T}{2}}+x_{\frac{T}{2}+1}+...+x_{T} \leq \frac{T}{2} \nonumber\\
    & 0 \leq x_t \leq 1\ \text{ for } t=1,...,T.\nonumber
\end{align}
For any online policy $\pi$, denote by $x^1_t(\pi)$ the decision of the policy $\pi$ at period $t$ under the true scenario given in \eqref{appendixeg3} and denote by $x^2_t(\pi)$ the decision of the policy $\pi$ at period $t$ under the true scenario \eqref{appendixeg4}. Further define $T_1(\pi)$ (resp. $T_2(\pi)$) as the expected capacity consumption of policy $\pi$ under the true scenario \eqref{appendixeg3} (resp. true scenario \eqref{appendixeg4}) during the first $\frac{T}{2}$ time periods:
\[
T_1(\pi)=\mathbb{E}\left[\sum_{t=1}^{\frac{T}{2}}x^1_t(\pi)\right] \text{~~~and~~~} T_2(\pi)=\mathbb{E}\left[\sum_{t=1}^{\frac{T}{2}}x^2_t(\pi)\right]
\]
Then, we have that
\[
\ALG_T^1(\pi)=-\frac{T+W_T}{2}+\frac{W_T}{T}\cdot T_1(\pi)\text{~~~and~~~}\ALG_T^2(\pi)=-\frac{T-W_T}{2}-\frac{W_T}{T}\cdot T_2(\pi)
\]
where $\ALG_T^1(\pi)$ (resp. $\ALG_T^2(\pi)$) denotes the expected reward collected by policy $\pi$ on scenario \eqref{appendixeg3} (resp. scenario \eqref{appendixeg4}). It is clear to see that the optimal policy $\pi^*$ who is aware of $P_t$ for each $t\in[T]$ can achieve an objective value
\[
\ALG_T^1(\pi^*)=-\frac{T+W_T}{2}\text{~~~and~~~}\ALG_T^2(\pi^*)=-\frac{T}{2}.
\]
Thus, the regret of policy $\pi$ on scenario \eqref{appendixeg3} and \eqref{appendixeg4} are $\frac{W_T}{T}\cdot T_1(\pi)$ and $W_T-\frac{W_T}{T}\cdot T_2(\pi)$ respectively. Further note that since the implementation of policy $\pi$ at each time period should be independent of future realizations, we must have $T_1(\pi)=T_2(\pi)$ (during the first $\frac{T}{2}$ periods, the information for $\pi$ is the same for both scenarios \eqref{appendixeg3} and \eqref{appendixeg4}). Thus, we have that
\[
\text{Reg}_T(\pi)\geq\max\left\{\frac{W_T}{T}\cdot T_1(\pi),W_T-\frac{W_T}{T}\cdot T_1(\pi) \right\}\geq\frac{W_T}{2}=\Omega(W_T)
\]
which completes our proof.
\end{myproof}

\begin{myproof}[Proof of \Cref{lem:decompose}]
Denote by $(\hat{\bm{\lambda}}^*, (\hbc^*_t)_{t=1}^T)$ the optimal solution to
\[
\max_{\bm{\lambda}\geq0}\min_{\bm{c}_t\in\mathcal{C}} \hLInner(\bm{C}, \bm{\lambda}),
\]
used in the definition \eqref{eqn:121902}.
we now show that $(\hat{\bm{\lambda}}^*, \hbc^*_t)$ is an optimal solution to
\[
\max_{\bm{\lambda}\geq0}\min_{\bm{c}_t\in\mathcal{C}} \hLInner_t(\bm{c}_t, \bm{\lambda})
\]
for each $t\in[T]$, which would help to complete our proof of \eqref{eqn:121904}. We first define
\[
L_t(\bm{\lambda})=\min_{\bm{c}_t\in\mathcal{C}} \hLInner_t(\bm{c}_t, \bm{\lambda}),
\]
as a function over $\bm{\lambda}$ for each $t\in[T]$. Then, it holds that
\begin{equation}\label{eqn:121905}
\nabla L_t(\hat{\bm{\lambda}}^*)=\nabla \hLInner_t(\hbc^*_t, \hat{\bm{\lambda}}^*)=\left( -\frac{\hat{\beta}_{i,t}}{T\beta_i}+\mathbb{E}_{\theta\sim\hat{P}_t}\left[ \frac{g_{i,\theta}(\hbc^*_t, \hbx^*_t(\theta))}{T\beta_i} \right] \right)_{\forall i\in[m]}=\bm{0}
\end{equation}
The first equality of \eqref{eqn:121905} follows from the fact that
\begin{equation}\label{eqn:121906}
\hbc^*_t\in\text{argmin}_{\bc\in\mathcal{C}}\mathbb{E}_{\theta\sim\hat{P}_t}\left[p(\bc)+
\min_{\bx\in\mathcal{K}(\theta,\bc)}f_{\theta}(\bx)+\sum_{i=1}^{m}\frac{\hat{\lambda}^*_i\cdot g_{i,\theta}(\bc, \bx)}{T\cdot\beta_i}\right]
\end{equation}
since $(\hat{\bm{\lambda}}^*, (\hbc^*_t)_{t=1}^T)$ is an optimal solution to $\max_{\bm{\lambda}\geq0}\min_{\bm{c}_t\in\mathcal{C}} \hLInner(\bm{C}, \bm{\lambda})$. The last equality of \eqref{eqn:121905} follows from the definition of $\hat{\beta}_{i,t}$ in \eqref{eqn:121902}. Therefore, combining \eqref{eqn:121905} and \eqref{eqn:121906}, we know that $(\hat{\bm{\lambda}}^*, \hbc^*_t)$ is an optimal solution to
$\max_{\bm{\lambda}\geq0}\min_{\bm{c}_t\in\mathcal{C}} \hLInner_t(\bm{c}_t, \bm{\lambda})$
for each $t\in[T]$.

We now prove \eqref{eqn:121904}. It is sufficient to prove that
\begin{equation}\label{eqn:121907}
  \hLInner(\hat{\bm{C}}^*, \hat{\bm{\lambda}}^*)=\sum_{t=1}^{T} \hLInner_t(\hbc^*_t, \hat{\bm{\lambda}}^*)
\end{equation}
where $\hat{\bm{C}}^*=(\hbc^*_t)_{t=1}^T$. We define an index set $\mathcal{I}=\{i\in[m]: \hat{\lambda}^*_i>0\}$. Clearly, for each $i\in\mathcal{I}$, the optimality of $\hat{\bm{\lambda}}^*$ would require that
\[
\nabla_{\lambda_i}\bLInner(\hat{\bm{C}}^*, \hat{\bm{\lambda}}^*)=-1+\frac{1}{T\beta_i}\sum_{t=1}^{T}\mathbb{E}_{\theta\sim\hat{P}_t}\left[ g_{i,\theta}(\hbc_t^*, \hbx^*_t(\theta)) \right]=0
\]
which implies $\sum_{t=1}^{T}\hat{\beta}_{i,t}=T\cdot\beta$. Therefore, we would have
\[
\sum_{i=1}^{m}\sum_{t=1}^{T}\frac{\hat{\lambda}^*_{i,t}\hat{\beta}_i}{T\beta_i}=\sum_{i=1}^{m}\hat{\lambda}^*_i.
\]
which completes our proof of \eqref{eqn:121908} and therefore \eqref{eqn:121904}.
\end{myproof}

\begin{myproof}[Proof of \Cref{thm:NStaRegretISP}]
The proof can be classified by the following two steps. We denote by
\begin{equation}\label{eqn:121909}
\LInner_t(\bm{c}, \bm{\lambda}):=\mathbb{E}_{\theta\sim P_t}\left[ p(\bc)-\sum_{i=1}^{m}\frac{\lambda_i\cdot\hat{\beta}_{i,t}}{T\cdot\beta_i}+
\min_{\bx\in\mathcal{K}(\theta,\bc)}f_{\theta}(\bx)+\sum_{i=1}^{m}\frac{\lambda_i\cdot g_{i,\theta}(\bc, \bx)}{T\cdot\beta_i} \right]
\end{equation}
for each $t\in[T]$. Our first step is to show that for any $\bm{\lambda}\geq0$ and any $\bm{c}\in\mathcal{C}$, it holds that
\begin{equation}\label{eqn:121910}
  \left|\LInner_t(\bm{c}, \bm{\lambda})-\hLInner_t(\bm{c}, \bm{\lambda})\right|\leq \frac{\|\bm{\lambda}\|_1}{T\beta_{\min}}\cdot W(\hat{P}_t, P_t)
\end{equation}
where $\beta_{\min}=\min_{i\in[m]}\{\beta_i\}$.

We now prove \eqref{eqn:121910}. We define
\[
\hLInner_t(\bm{c}, \bm{\lambda}, \theta):= p(\bc)-\sum_{i=1}^{m}\frac{\lambda_i\cdot\hat{\beta}_{i,t}}{T\cdot\beta_i}+
\min_{\bx\in\mathcal{K}(\theta,\bc)}f_{\theta}(\bx)+\sum_{i=1}^{m}\frac{\lambda_i\cdot g_{i,\theta}(\bc, \bx)}{T\cdot\beta_i}.
\]
It is clear to see that
\[
\LInner_t(\bm{c}, \bm{\lambda})=\mathbb{E}_{\theta\sim P_t}\left[ \hLInner_t(\bm{c}, \bm{\lambda}, \theta) \right]\text{~and~}\hLInner_t(\bm{c}, \bm{\lambda})=\mathbb{E}_{\theta\sim \hat{P}_t}\left[ \hLInner_t(\bm{c}, \bm{\lambda}, \theta) \right].
\]
Moreover, note that for any $\theta, \theta'$, we have
\[
|\hLInner_t(\bm{c}, \bm{\lambda}, \theta)-\hLInner_t(\bm{c}, \bm{\lambda}, \theta')|\leq \frac{\|\bm{\lambda}\|_1}{T\beta_{\min}}\cdot d(\theta, \theta'),
\]
with $d(\theta, \theta')=\|(f_{\theta}, \bm{g}_{\theta})-(f_{\theta'}, \bm{g}_{\theta'})\|_{\infty}$ in the definition \eqref{eqn:Wasserdist}. Therefore, we have
\[
\left|\LInner_t(\bm{c}, \bm{\lambda})-\hLInner_t(\bm{c}, \bm{\lambda})\right|=\left| \mathbb{E}_{\theta\sim P_t}\left[\hLInner_t(\bm{c}, \bm{\lambda}, \theta)\right]-\mathbb{E}_{\theta'\sim\hat{P}_t}\left[\hLInner_t(\bm{c}, \bm{\lambda}, \theta')\right] \right|\leq \frac{\|\bm{\lambda}\|_1}{T\beta_{\min}}\cdot W(\hat{P}_t, P_t),
\]
thus, complete our proof of \eqref{eqn:121910}.

Our second step is to bound the final regret with the help of \eqref{eqn:121910}. We assume without loss of generality that there always exists $i'\in[m]$ such that
\begin{equation}\label{eqn:121915}
\sum_{t=1}^{T} g_{i', \theta_t}(\bc_t, \bm{x}_t)\geq T\cdot \beta_{i'}.
\end{equation}
In fact, let there be a \textit{dummy} constraint $i'$ such that $g_{i',\theta}(\bc, \bm{x})=\beta_{i'}=\alpha$, for arbitrary $\alpha\in(0,1)$, for any $\theta$ and $\bc, \bm{x}$. Then, \eqref{eqn:121915} holds.

Let $(\hat{\bm{\lambda}}^*, (\hbc^*_t)_{t=1}^T)$ be the optimal solution to $\max_{\bm{\lambda}\geq0}\min_{\bm{c}_t\in\mathcal{C}} \hLInner(\bm{C}, \bm{\lambda})$ used in the definition \eqref{eqn:121902}. Then, it holds that
\begin{align}
\mathbb{E}_{\bm{\theta}\sim\bm{P}}\left[ \sum_{t=1}^{T}\hLInner_t(\bm{c}_t, \mu\cdot\bm{e}_{i_t}, \theta_t) \right]&=\sum_{t=1}^{T}\mathbb{E}_{\bc_t, i_t}\left[\LInner_t(\bc_t, \mu\cdot\bm{e}_{i_t})\right]\label{eqn:121911}\\
&\leq\sum_{t=1}^{T}\mathbb{E}_{\bc_t, i_t}\left[\hLInner_t(\bm{c}_t, \mu\cdot\bm{e}_{i_t})\right]+\frac{\mu\cdot W_T}{T\beta_{\min}}=\sum_{t=1}^{T}\mathbb{E}_{i_t}\left[ \min_{\bc\in\mathcal{C}}\hLInner_t(\bm{c}, \mu\cdot\bm{e}_{i_t}) \right]+\frac{\mu\cdot W_T}{T\beta_{\min}}\nonumber\\
&\leq \sum_{t=1}^{T} \min_{\bc\in\mathcal{C}}\hLInner_t(\bm{c}, \hat{\bm{\lambda}}^*)+\frac{\mu\cdot W_T}{T\beta_{\min}}=\sum_{t=1}^{T} \hLInner_t(\hbc^*_t, \hat{\bm{\lambda}}^*)+\frac{\mu\cdot W_T}{T\beta_{\min}} \nonumber
\end{align}
where the first inequality follows from the definition of $\bc_t$, the second inequality follows from \Cref{lem:decompose}, and the last equality follows from the definition of $\hbc^*_t$.

On the other hand, for any $i\in[m]$, we have
\[
\sum_{t=1}^{T}\hLInner_t(\bm{c}_t, \mu\cdot\bm{e}_{i_t}, \theta_t)\geq \sum_{t=1}^{T}\hLInner_{i,t}(\bm{c}_t, \mu\cdot\bm{e}_{i_t}, \theta_t)-\text{Reg}(T, \bm{\theta})
\]
with $\hLInner_{i,t}$ defined in \eqref{eqn:121916}, where $\text{Reg}(\tau, \bm{\theta})$ denotes the regret bound of $\ALGD$ (holds for arbitrary $\bm{\lambda}=\mu\cdot\bm{e}_i$). We now denote by
\[
i^*=\text{argmax}_{i\in[m]}\{ \frac{1}{T}\cdot\sum_{t=1}^{T}g_{i,\theta_t}(\bc_t, \bm{x}_t)-\beta_i \}.
\]
We also denote by
\[
d_T(\mathcal{A}, \bm{\theta})=\max_{i\in[m]}\{ \frac{1}{T}\cdot\sum_{t=1}^{T}g_{i,\theta_t}(\bc_t, \bm{x}_t)-\beta_i \}.
\]
From \eqref{eqn:121915}, we must have $d_T(\mathcal{A}, \bm{\theta})\geq0$. We now set $i=i^*$ and we have
\begin{align}
\sum_{t=1}^{T}\hLInner_t(\bm{c}_t, \mu\cdot\bm{e}_{i_t}, \theta_t)&\geq \sum_{t=1}^{T}\hLInner_{i^*,t}(\bm{c}_t, \mu\cdot\bm{e}_{i_t}, \theta_t)-\text{Reg}(T, \bm{\theta}) \label{eqn:122201}\\
&=\sum_{t=1}^{T} \left(p(\bm{c}_t)+f_{\theta_t}(\bm{x}_t)\right)-\mu\cdot \frac{\sum_{t=1}^{T}\hat{\beta}_{i^*,t}}{T\beta_{i^*}}+\sum_{t=1}^{T}\frac{\mu\cdot g_{i^*,\theta_t}(\bc_t, \bm{x}_t)}{T\cdot\beta_{i^*}}-\text{Reg}(T, \bm{\theta}) \nonumber
\end{align}
From the construction of $\hat{\beta}_{i,t}$, we know that $\sum_{t=1}^{T}\hat{\beta}_{i,t}\leq T\cdot\beta_i$ for each $i\in[m]$. To see this point, we note that if we define $\hat{L}(\bm{\lambda})=\min_{\bm{c}_t\in\mathcal{C}} \hLInner(\bm{C}, \bm{\lambda})$, then, for each $i\in[m]$,
\begin{equation}\label{eqn:122202}
\nabla_{\lambda_i} \hat{L}(\hat{\bm{\lambda}}^*)=-1+\mathbb{E}_{\bm{\theta}\sim\hat{P}}\left[\sum_{t=1}^{T}\frac{g_{i,\theta_t}(\hbc^*_t, \hbx^*(\theta_t))}{T\beta_i}\right]
=-1+\frac{\sum_{t=1}^{T}\hat{\beta}_{i,t}}{T\cdot\beta_i}\leq 0.
\end{equation}
Otherwise, $\nabla_{\lambda_i} \hat{L}(\hat{\bm{\lambda}}^*)>0$ would violate the optimality of $\hat{\bm{\lambda}}^*$ to $\max_{\bm{\lambda}\geq0} \hat{L}(\bm{\lambda})$.

Plugging \eqref{eqn:122202} into \eqref{eqn:122201}, we get
\begin{align}
\sum_{t=1}^{T}\hLInner_t(\bm{c}_t, \mu\cdot\bm{e}_{i_t}, \theta_t)&\geq \sum_{t=1}^{T} \left(p(\bm{c}_t)+f_{\theta_t}(\bm{x}_t)\right)-\mu+\sum_{t=1}^{T}\frac{\mu\cdot g_{i^*,\theta_t}(\bc_t, \bm{x}_t)}{T\cdot\beta_{i^*}}-\text{Reg}(T, \bm{\theta})\label{eqn:121912}\\
&\geq \sum_{t=1}^{T} \left(p(\bm{c}_t)+f_{\theta_t}(\bm{x}_t)\right)+\frac{\mu}{\beta_{\min}}\cdot d_T(\mathcal{A}, \bm{\theta})-\text{Reg}(T, \bm{\theta})\nonumber
\end{align}
Combining \eqref{eqn:121911} and \eqref{eqn:121912}, we have
\begin{equation}\label{eqn:121918}
\mathbb{E}_{\bm{\theta}\sim\bm{P}}\left[ \sum_{t=1}^{T} \left(p(\bm{c}_t)+f_{\theta_t}(\bm{x}_t)\right) \right]\leq \sum_{t=1}^{T} \hLInner_t(\hbc^*_t, \hat{\bm{\lambda}}^*)+\frac{\mu\cdot W_T}{T\beta_{\min}}+\mathbb{E}_{\bm{\theta}\sim\bm{P}}\left[\text{Reg}(T, \bm{\theta})\right]-\frac{\mu}{\beta_{\min}}\cdot\mathbb{E}_{\bm{\theta}\sim\bm{P}}\left[ d_T(\mathcal{A}, \bm{\theta}) \right].
\end{equation}
Denote by $(\bm{\lambda}^*, \tilde{\bm{C}}^*)$ the optimal \textit{saddle-point} solution to
\[
\max_{\bm{\lambda}\geq0} \min_{\tilde{\bm{C}}\in\mathcal{C}} \mathbb{E}_{\tilde{\bm{C}}}[\LInner(\tilde{\bm{C}}, \bm{\lambda})]
=\min_{\tilde{\bm{C}}\in\mathcal{C}}\max_{\bm{\lambda}\geq0}\mathbb{E}_{\tilde{\bm{C}}}[\LInner(\tilde{\bm{C}}, \bm{\lambda})].
\]
Then, it holds that
\begin{align}
\sum_{t=1}^{T} \hLInner_t(\hbc^*_t, \hat{\bm{\lambda}}^*)&\leq \sum_{t=1}^{T} \mathbb{E}_{\tbc_t}[\hLInner_t(\tbc^*_t, \hat{\bm{\lambda}}^*)]\leq \sum_{t=1}^{T} \mathbb{E}_{\tbc_t}[\LInner_t(\tbc^*_t, \hat{\bm{\lambda}}^*)]+\frac{\|\hat{\bm{\lambda}}^*\|_1\cdot W_T}{T\beta_{\min}}=\mathbb{E}_{\tilde{\bm{C}}}[\LInner(\tilde{\bm{C}}^*, \hat{\bm{\lambda}}^*)]+\frac{\|\hat{\bm{\lambda}}^*\|_1\cdot W_T}{T\beta_{\min}}  \nonumber\\
&\leq \mathbb{E}_{\tilde{\bm{C}}}[\LInner(\tilde{\bm{C}}^*, \hat{\bm{\lambda}}^*)]+\frac{\|\hat{\bm{\lambda}}^*\|_1\cdot W_T}{T\beta_{\min}}=\OPTInner+\frac{\|\hat{\bm{\lambda}}^*\|_1\cdot W_T}{T\beta_{\min}}, \label{eqn:122003}
\end{align}
where the first inequality follows from definition of $(\hat{\bm{\lambda}^*}, (\hbc_t)_{t=1}^T)$, the second inequality follows from \eqref{eqn:121910}, the first equality follows from \eqref{eqn:121908} and the third inequality follows from the saddle-point condition of $(\bm{\lambda}^*, \tilde{\bm{C}}^*)$.

Plugging \eqref{eqn:122003} into \eqref{eqn:121918}, we have
\begin{equation}\label{eqn:122004}
\mathbb{E}_{\bm{\theta}\sim\bm{P}}\left[ \sum_{t=1}^{T} \left(p(\bm{c}_t)+f_{\theta_t}(\bm{x}_t)\right) \right]\leq \OPTInner+\frac{2\mu W_T}{T\beta_{\min}}+\mathbb{E}_{\bm{\theta}\sim\bm{P}}\left[\text{Reg}(T, \bm{\theta})\right]-\frac{\mu}{\beta_{\min}}\cdot\mathbb{E}_{\bm{\theta}\sim\bm{P}}\left[ d_T(\mathcal{A}, \bm{\theta}) \right].
\end{equation}
with $\mu=\|\hat{\bm{\lambda}}^*\|_1$.
From the non-negativity of $d_T(\mathcal{A}, \bm{\theta})$, we have
\[
\mathbb{E}_{\bm{\theta}\sim\bm{P}}\left[ \sum_{t=1}^{T} \left(p(\bm{c}_t)+f_{\theta_t}(\bm{x}_t)\right) \right]\leq \OPT+\frac{2\mu W_T}{T\beta_{\min}}+\mathbb{E}_{\bm{\theta}\sim\bm{P}}\left[\text{Reg}(T, \bm{\theta})\right].
\]
Using \Cref{thm:Hedgeregret} to bound $\mathbb{E}_{\bm{\theta}\sim\bm{P}}\left[\text{Reg}(T, \bm{\theta})\right]$, we have
\[
\mathbb{E}_{\bm{\theta}\sim\bm{P}}\left[ \sum_{t=1}^{T} \left(p(\bm{c}_t)+f_{\theta_t}(\bm{x}_t)\right) \right]\leq \OPT+\frac{2\mu W_T}{T\beta_{\min}}+\tilde{O}(\sqrt{T\cdot\log m})
\]
which completes our proof of \eqref{eqn:122001} by noting that $\mu=\alpha\cdot T$ for some constant $\alpha>0$.

We note that $(\bm{c}_t, \bm{x}_t)_{t=1}^T$ defines a feasible solution to
\begin{align}
 \OPT^{\delta}= \min & \sum_{t=1}^{T}\mathbb{E}_{\bm{c}_t, \bm{x}_t, \theta_t\sim P_t}\left[p(\bm{c}_t)+f_{\theta_t}(\bm{x}_t)\right] \label{lp:NOuterA}\\
  \mbox{s.t.} & \frac{1}{T}\cdot\sum_{t=1}^{T}\mathbb{E}_{\bc_t, \bm{x}_t, \theta_t\sim P_t}[\bm{g}_{\theta_t}(\bc_t, \bm{x}_t)]\leq\bm{\beta}+\delta\cdot\bm{e} \nonumber\\
  & \bm{x}_t\in\mathcal{K}(\theta_t, \bm{c}_t), \bm{c}_t\in\mathcal{C}, \forall t.\nonumber
\end{align}
with $\delta=\mathbb{E}_{\bm{\theta}\sim\bm{P}}[d_T(\mathcal{A}, \bm{\theta})]$ and $\bm{e}$ denotes a vector of all ones. We define another optimization problem by changing $P_t$ into $\hat{P}_t$ for each $t$,
\begin{align}
 \hOPT^{\hat{\delta}}= \min & \sum_{t=1}^{T}\mathbb{E}_{\bm{c}_t, \bm{x}_t, \theta_t\sim\hat{P}_t}\left[p(\bm{c}_t)+f_{\theta_t}(\bm{x}_t)\right] \label{lp:hNOuterA}\\
  \mbox{s.t.} & \frac{1}{T}\cdot\sum_{t=1}^{T}\mathbb{E}_{\bc_t, \bm{x}_t, \theta_t\sim\hat{P}_t}[\bm{g}_{\theta_t}(\bc_t, \bm{x}_t)]\leq\bm{\beta}+\hat{\delta}\cdot\bm{e} \nonumber\\
  & \bm{x}_t\in\mathcal{K}(\theta_t, \bm{c}_t), \bm{c}_t\in\mathcal{C}, \forall t.\nonumber
\end{align}
We have the following result that bounds the gap between $\OPT^{\delta}$ and $\hOPT^{\hat{\delta}}$, for some specific $\delta$ and $\hat{\delta}$.
\begin{claim}\label{claim:Bound}
For any $\delta\geq0$, if we set $\hat{\delta}=\delta+\frac{W_T}{T}$, then we have
\[
\hOPT^{\hat{\delta}}\leq\OPT^{\delta}+W_T.
\]
\end{claim}

If we regard $\hOPT^{\hat{\delta}}$ as a function over $\hat{\delta}$, then $\hOPT^{\hat{\delta}}$ is clearly a convex function over $\hat{\delta}$, where the proof follows the same spirit as the proof of \Cref{lem:ConvexL}. Moreover, note that
\[
\frac{d\hOPT^{\hat{\delta}=0}}{d\hat{\delta}}=\|\hat{\bm{\lambda}}^*\|_1\leq\mu.
\]
We have
\[
\hOPT=\hOPT^{0}\leq \hOPT^{\hat{\delta}}+  \mu\cdot \hat{\delta}
\]
for any $\hat{\delta}\geq0$. On the other hand, we have
\begin{equation}\label{eqn:010401}
\hOPT=\max_{\bm{\lambda}\geq0}\min_{\bm{c}_t\in\mathcal{C}} \hLInner(\bm{C}, \bm{\lambda})=\sum_{t=1}^{T} \hLInner_t(\hbc^*_t, \hat{\bm{\lambda}}^*)
\end{equation}
where the last equality follows from \Cref{lem:decompose}. Therefore, we now assume that $\mathbb{E}_{\bm{\theta}\sim\bm{P}}[d_T(\mathcal{A}, \bm{\theta})]\geq\frac{W_T}{T}$, otherwise \eqref{eqn:122002} directly holds. We have
\begin{equation}\label{eqn:121914}
\begin{aligned}
\sum_{t=1}^{T} \hLInner_t(\hbc^*_t, \hat{\bm{\lambda}}^*)&=\hOPT \leq \hOPT^{\mathbb{E}_{\bm{\theta}\sim\bm{P}}[d_T(\mathcal{A},\bm{\theta})]+\frac{W_T}{T}}+ \mu\cdot \left(\mathbb{E}_{\bm{\theta}\sim\bm{P}}[d_T(\mathcal{A},\bm{\theta})]+\frac{W_T}{T}\right)\\
&\leq \OPT^{\mathbb{E}_{\bm{\theta}\sim\bm{P}}[d_T(\mathcal{A},\bm{\theta})]}+ \mu\cdot \mathbb{E}_{\bm{\theta}\sim\bm{P}}[d_T(\mathcal{A},\bm{\theta})]+W_T\cdot(1+\frac{\mu}{T})\\
&\leq \sum_{t=1}^{T} \mathbb{E}_{\bm{\theta}\sim\bm{P}}\left[p(\bm{c}_t)+f_{\theta_t}(\bm{x}_t)\right]+\mu\cdot \mathbb{E}_{\bm{\theta}\sim\bm{P}}[d_T(\mathcal{A},\bm{\theta})]+W_T\cdot(1+\frac{\mu}{T})
\end{aligned}
\end{equation}
where the second inequality follows from \Cref{claim:Bound} by setting $\delta=\mathbb{E}_{\bm{\theta}\sim\bm{P}}[d_T(\mathcal{A},\bm{\theta})]$.
Plugging \eqref{eqn:121914} into \eqref{eqn:121918}, we have
\[\begin{aligned}
\sum_{t=1}^{T} \hLInner_t(\hbc^*_t, \hat{\bm{\lambda}}^*)\leq & \sum_{t=1}^{T} \hLInner_t(\hbc^*_t, \hat{\bm{\lambda}}^*)+\frac{\mu\cdot W_T}{T\beta_{\min}}+\mathbb{E}_{\bm{\theta}\sim\bm{P}}\left[\text{Reg}(T, \bm{\theta})\right]-\frac{\mu}{\beta_{\min}}\cdot\mathbb{E}_{\bm{\theta}\sim\bm{P}}\left[ d_T(\mathcal{A}, \bm{\theta}) \right]\\
&+\mu\cdot \mathbb{E}_{\bm{\theta}\sim\bm{P}}[d_T(\mathcal{A},\bm{\theta})]+W_T\cdot(1+\frac{\mu}{T}),
\end{aligned}\]
which implies
\[
\mu\cdot\left(\frac{1}{\beta_{\min}}-1\right)\cdot \mathbb{E}_{\bm{\theta}\sim\bm{P}}[ d_T(\mathcal{A},\bm{\theta})]\leq \mathbb{E}_{\bm{\theta}\sim\bm{P}}\left[\text{Reg}(T, \bm{\theta})\right]+\frac{2\mu W_T}{T\beta_{\min}}++W_T\cdot(1+\frac{\mu}{T}).
\]
which completes our final proof by noting that $\mu=\alpha\cdot T$ for some constant $\alpha>0$.
\end{myproof}

\begin{myproof}[Proof of \Cref{claim:Bound}]
Denote by $(\tilde{\bm{c}}_t, \tilde{\bm{x}}_t)_{t=1}^T$ one optimal solution to $\OPT^{\delta}$. Then, from the definition of $W_T$, we have that
\[
\frac{1}{T}\cdot\sum_{t=1}^{T}\mathbb{E}_{\tilde{\bc}_t, \tilde{\bm{x}}_t, \theta_t\sim\hat{P}_t}[\bm{g}_{\theta_t}(\bc_t, \bm{x}_t)]\leq \frac{1}{T}\cdot\sum_{t=1}^{T}\mathbb{E}_{\tilde{\bc}_t, \tilde{\bm{x}}_t, \theta_t\sim P_t}[\bm{g}_{\theta_t}(\bc_t, \bm{x}_t)]+\frac{W_T}{T}.
\]
Therefore, we know that $(\tilde{\bm{c}}_t, \tilde{\bm{x}}_t)_{t=1}^T$ is a feasible solution to $\hOPT^{\hat{\delta}}$. On the other hand, from the definition of $W_T$, we know that
\[
\hOPT^{\hat{\delta}}\leq\sum_{t=1}^{T}\mathbb{E}_{\tilde{\bm{c}}_t, \tilde{\bm{x}}_t, \theta_t\sim\hat{P}_t}\left[p(\tilde{\bm{c}}_t)+f_{\theta_t}(\tilde{\bm{x}}_t)\right]\leq \sum_{t=1}^{T}\mathbb{E}_{\tilde{\bm{c}}_t, \tilde{\bm{x}}_t, \theta_t\sim\hat{P}_t}\left[p(\tilde{\bm{c}}_t)+f_{\theta_t}(\tilde{\bm{x}}_t)\right]+W_T=\OPT^{\delta}+W_T
\]
by noting that the distribution of $\tilde{\bm{c}}_t$ must be independent of $\theta_t$ for each $t$, which completes our proof.
\end{myproof}

\section{Missing Proofs for Section \ref{sec:stationary}}
\begin{myproof}[Proof of \Cref{thm:adverLower}]
The proof is modified from the proof of Theorem 2 in \cite{jiang2020online}.
We let $W_T=\mathbb{E}_{\bm{\theta}\sim\bm{P}}[W(\bm{\theta})]$.
We consider a special case of our problem where for any $\bc\in\mathcal{C}$ and any $\theta$, $\mathcal{K}(\theta, \bc)=[0,1]$ (there is no need to decide the first-stage decision). There is only one long-term constraint with target $\beta=\frac{1}{2}$. Moreover, there are three possible values of $\theta$, denoted by $\{\theta^1, \theta^2, \theta^3\}$. We have $f_{\theta^1}(x)=-x$, $f_{\theta^2}(x)=-\left(1+\frac{W_T}{T}\right)x$, $f_{\theta^3}(x)=-\left(1-\frac{W_T}{T}\right)x$ and $g_{\theta^1}(x)=g_{\theta^2}(x)=g_{\theta^3}(x)=x$ (only one long-term constraint). The true distribution is $P_t=\theta^1$ with probability 1 for each $t\in[T]$, and the problem with respect to the true distributions can be described below in \eqref{newappendixeg0}.
\begin{align}
   \min \ \ &  -x_1-...-x_{\frac{T}{2}}-x_{\frac{T}{2}+1}-...-x_{T}  \label{newappendixeg0} \\
    \text{s.t. }\ & x_1+...+x_{\frac{T}{2}}+x_{\frac{T}{2}+1}+...+x_{T} \leq \frac{T}{2}\nonumber \\
    & 0 \leq x_t \leq 1\ \text{ for } t=1,...,T. \nonumber
\end{align}
Now we consider the following two possible adversarial corruptions. The first possible corruption, given in \eqref{newappendixeg3}, is that the distribution $P^c_t=\theta^2$ for $t=\frac{T}{2}+1,\dots,T$. The second possible corruption, given in \eqref{newappendixeg4}, is that the distribution $P^c_t=\theta^3$ for $t=\frac{T}{2}+1,\dots,T$.
\begin{align}
   \min \ \ &  -x_1-...-x_{\frac{T}{2}}-\left(1+\frac{W_T}{T}\right)x_{\frac{T}{2}+1}-...-\left(1+\frac{W_T}{T}\right)x_{T}  \label{newappendixeg3} \\
    \text{s.t. }\ & x_1+...+x_{\frac{T}{2}}+x_{\frac{T}{2}+1}+...+x_{T} \le \frac{T}{2}\nonumber \\
    & 0 \leq x_t \leq 1\ \text{ for } t=1,...,T. \nonumber \\
   \min \ \ &  -x_1-...-x_{\frac{T}{2}} -\left(1-\frac{W_T}{T}\right)x_{\frac{T}{2}+1}-...-\left(1-\frac{W_T}{T}\right)x_{T} \label{newappendixeg4} \\
    \text{s.t. }\ & x_1+...+x_{\frac{T}{2}}+x_{\frac{T}{2}+1}+...+x_{T} \leq \frac{T}{2} \nonumber\\
    & 0 \leq x_t \leq 1\ \text{ for } t=1,...,T.\nonumber
\end{align}
For any online policy $\pi$, denote by $x^1_t(\pi)$ the decision of the policy $\pi$ at period $t$ under corruption scenario given in \eqref{newappendixeg3} and denote by $x^2_t(\pi)$ the decision of the policy $\pi$ at period $t$ under corruption scenario \eqref{newappendixeg4}. Further define $T_1(\pi)$ (resp. $T_2(\pi)$) as the expected capacity consumption of policy $\pi$ under corruption scenario \eqref{newappendixeg3} (resp. corruption scenario \eqref{newappendixeg4}) during the first $\frac{T}{2}$ time periods:
\[
T_1(\pi)=\mathbb{E}\left[\sum_{t=1}^{\frac{T}{2}}x^1_t(\pi)\right] \text{~~~and~~~} T_2(\pi)=\mathbb{E}\left[\sum_{t=1}^{\frac{T}{2}}x^2_t(\pi)\right]
\]
Then, we have that
\[
\ALG_T^1(\pi)=-\frac{T+W_T}{2}+\frac{W_T}{T}\cdot T_1(\pi)\text{~~~and~~~}\ALG_T^2(\pi)=-\frac{T-W_T}{2}-\frac{W_T}{T}\cdot T_2(\pi)
\]
where $\ALG_T^1(\pi)$ (resp. $\ALG_T^2(\pi)$) denotes the expected reward collected by policy $\pi$ on scenario \eqref{newappendixeg3} (resp. scenario \eqref{newappendixeg4}). It is clear to see that the optimal policy $\pi^*$ who is aware of $P^c_t$ for each $t\in[T]$ can achieve an objective value
\[
\ALG_T^1(\pi^*)=-\frac{T+W_T}{2}\text{~~~and~~~}\ALG_T^2(\pi^*)=-\frac{T}{2}.
\]
Thus, the regret of policy $\pi$ on scenario \eqref{newappendixeg3} and \eqref{newappendixeg4} are $\frac{W_T}{T}\cdot T_1(\pi)$ and $W_T-\frac{W_T}{T}\cdot T_2(\pi)$ respectively. Further note that since the implementation of policy $\pi$ at each time period should be independent of future realizations, and more importantly, should independent of corruptions in the future, we must have $T_1(\pi)=T_2(\pi)$ (during the first $\frac{T}{2}$ periods, the information for $\pi$ is the same for both scenarios \eqref{newappendixeg3} and \eqref{newappendixeg4}). Thus, we have that
\[
\text{Reg}_T(\pi)\geq\max\left\{\frac{W_T}{T}\cdot T_1(\pi),W_T-\frac{W_T}{T}\cdot T_1(\pi) \right\}\geq\frac{W_T}{2}=\Omega(W_T)
\]
which completes our proof.
\end{myproof}

\begin{myproof}[Proof of \Cref{lem:ConvexL}]
Fix arbitrary $\bm{\lambda}, \theta$, for any $\bm{c}_1, \bm{c}_2\in\mathcal{C}$ and any $\alpha_1, \alpha_2\geq0$ such that $\alpha_1+\alpha_2=1$, we prove
\[
\alpha_1\cdot\bLInner(\bm{c}_1,\bm{\lambda},\theta)+\alpha_2\cdot\bLInner(\bm{c}_2,\bm{\lambda},\theta)\geq \bLInner(\alpha_1\cdot\bm{c}_1+\alpha_2\cdot\bm{c}_2,\bm{\lambda},\theta).
\]
Now, for $\bLInner(\bm{c}_1,\bm{\lambda},\theta)$, we denote by $\bm{x}_1^*(\theta)$ one optimal solution of the inner minimization problem in the definition of $\bLInner(\bm{c}_1,\bm{\lambda},\theta)$ \eqref{eqn:LISPstationary}. Similarly, for $\bLInner(\bm{c}_2,\bm{\lambda},\theta)$, we denote by $\bm{x}_2^*(\theta)$ one optimal solution of the inner minimization problem in the definition of $\bLInner(\bm{c}_2,\bm{\lambda},\theta)$ \eqref{eqn:LISPstationary}. We then define
\[
\bm{x}^*_3(\theta)=\alpha_1\cdot\bm{x}^*_1(\theta)+\alpha_2\cdot\bm{x}^*_2(\theta),~~\forall \theta.
\]
Under \Cref{assump:main}, from the convexity of $f_{\theta}(\cdot)$ and $\bm{g}_{\theta}(\cdot)$, it holds that
\[\begin{aligned}
&\alpha_1\cdot\left(f_{\theta}(\bm{x}^*_1(\theta))+\sum_{i=1}^{m}\frac{\lambda_i\cdot g_{i,\theta}(\bc_1, \bm{x}^*_1(\theta))}{T\cdot\beta_i}\right)+\alpha_2\cdot\left(f_{\theta}(\bm{x}^*_2(\theta))+\sum_{i=1}^{m}\frac{\lambda_i\cdot g_{i,\theta}(\bc_2, \bm{x}^*_2(\theta))}{T\cdot\beta_i}\right)\\
&\geq f_{\theta}(\bm{x}^*_3(\theta))+\sum_{i=1}^{m}\frac{\lambda_i\cdot g_{i,\theta}(\bc_3, \bm{x}^*_3(\theta))}{T\cdot\beta_i}.
\end{aligned}\]
with $\bc_3=\alpha_1\cdot\bc_1+\alpha_2\cdot\bc_2$.
Given the convexity of $p(\cdot)$, we have
\[\begin{aligned}
&\alpha_1\cdot\bLInner(\bm{c}_1,\bm{\lambda},\theta)+\alpha_2\cdot\bLInner(\bm{c}_2,\bm{\lambda},\theta)\\
=&\alpha_1 p(\bm{c}_1)+\alpha_2 p(\bm{c}_2)-\frac{1}{T}\sum_{i=1}^{m}\lambda_i+\alpha_1\left(f_{\theta}(\bm{x}^*_1(\theta))+\sum_{i=1}^{m}\frac{\lambda_i\cdot g_{i,\theta}(\bc_1, \bm{x}^*_1(\theta))}{T\cdot\beta_i}\right)\\
&+\alpha_2\left(f_{\theta}(\bm{x}^*_2(\theta))+\sum_{i=1}^{m}\frac{\lambda_i\cdot g_{i,\theta}(\bc_2, \bm{x}^*_2(\theta))}{T\cdot\beta_i}\right)\\
\geq& p(\alpha_1\bm{c}_1+\alpha_2\bm{c}_2)+f_{\theta}(\bm{x}^*_3(\theta))+\sum_{i=1}^{m}\frac{\lambda_i\cdot g_{i,\theta}(\bc_3, \bm{x}^*_3(\theta))}{T\cdot\beta_i}\\
\geq& \bLInner(\bc_3,\bm{\lambda},\theta)
\end{aligned}\]
where the last inequality follows from $\bm{x}^*_3(\theta)\in\mathcal{K}(\theta,\bc_3)$, given the exact formulation of $\mathcal{K}(\theta, \bm{c})$ under \Cref{assump:main}.
\end{myproof}

\begin{myproof}[Proof of \Cref{thm:CorruptRegretISP}]
Denote by $\tau$ the time period that \Cref{alg:DAL} is terminated. There must be a constraint $i'\in[m]$ such that
\begin{equation}\label{eqn:121702}
\sum_{t=1}^{\tau} g_{i', \theta^c_t}(\bc_t, \bm{x}_t)\geq T\cdot \beta_{i'}.
\end{equation}
Otherwise, we can assume without loss of generality that there exists a \textit{dummy} constraint $i'$ such that $g_{i',\theta}(\bc_t, \bm{x})=\beta_{i'}=\alpha$, for arbitrary $\alpha\in(0,1)$, for any $\theta$ and $\bc, \bm{x}$. In this case, we can set $\tau=T$.

We denote by $(\bm{\lambda}^*, \tbc^*)$ one \textit{saddle-point} optimal solution to
\[
\max_{\bm{\lambda}\geq0}\min_{\tbc\in\mathcal{C}} \mathbb{E}_{\tbc^*, \theta\sim P^c}\left[\bLInner(\tbc, \bm{\lambda}, \theta)\right]=\min_{\tbc\in\mathcal{C}}\max_{\bm{\lambda}\geq0} \mathbb{E}_{\tbc^*, \theta\sim P^c}\left[\bLInner(\tbc, \bm{\lambda}, \theta)\right]
\]
where $P^c$ denotes the uniform mixture of the distributions of $\{\theta^c_t\}$ for $t=1$ to $\tau$ and the equality follows from the concavity over $\bm{\lambda}$ and the convexity over $\bm{c}$ proved in \Cref{lem:ConvexL}.
We have
\[
\sum_{t=1}^{\tau}\bLInner(\bm{c}_t, \mu\cdot\bm{e}_{i_t}, \theta^c_t)\leq \sum_{t=1}^{\tau}\mathbb{E}_{\tbc^*}\left[\bLInner(\tbc^*, \mu\cdot\bm{e}_{i_t}, \theta^c_t)\right]+\text{Reg}_1(\tau, \bm{\theta}^c)
\]
following regret bound of $\ALG_1$. Then, it holds that
\begin{align}
\mathbb{E}_{\bm{\theta}\sim\bm{P}}\left[ \sum_{t=1}^{\tau}\bLInner(\bm{c}_t, \mu\cdot\bm{e}_{i_t}, \theta^c_t) \right]&\leq \mathbb{E}_{\tbc^*, \bm{\theta}\sim\bm{P}}\left[ \sum_{t=1}^{\tau}\bLInner(\tbc^*, \mu\cdot\bm{e}_{i_t}, \theta^c_t) \right]+\mathbb{E}_{\bm{\theta}\sim\bm{P}}\left[\text{Reg}_1(\tau, \bm{\theta}^c)\right] \label{eqn:121701}\\
&\leq \tau\cdot \mathbb{E}_{\tbc^*, \theta\sim P^c}\left[ \bLInner(\tbc^*, \bm{\lambda}^*, \theta) \right]+\mathbb{E}_{\bm{\theta}\sim\bm{P}}\left[\text{Reg}_1(\tau, \bm{\theta}^c)\right] \nonumber\\
&\leq \tau\cdot \mathbb{E}_{\tbc^*, \theta\sim P}\left[ \bLInner(\tbc^*, \bm{\lambda}^*, \theta) \right]+\mathbb{E}_{\bm{\theta}\sim\bm{P}}\left[\text{Reg}_1(\tau, \bm{\theta}^c)\right]+O(\mathbb{E}_{\bm{\theta}\sim\bm{P}}[W(\bm{\theta})]). \nonumber
\end{align}
On the other hand, for any $i\in[m]$, we have
\[
\sum_{t=1}^{\tau}\bLInner(\bm{c}_t, \mu\cdot\bm{e}_{i_t}, \theta^c_t)\geq \sum_{t=1}^{\tau}\bLInner_i(\bm{c}_t, \mu\cdot\bm{e}_{i_t}, \theta^c_t)-\text{Reg}_2(\tau, \bm{\theta}^c)
\]
following the regret bound of $\ALG_2$ (holds for arbitrary $\bm{\lambda}=\mu\cdot\bm{e}_i$). We now set $i=i'$ and we have
\[\begin{aligned}
\sum_{t=1}^{\tau}\bLInner(\bm{c}_t, \mu\cdot\bm{e}_{i_t}, \theta^c_t)&\geq \sum_{t=1}^{\tau}\bLInner_{i'}(\bm{c}_t, \mu\cdot\bm{e}_{i_t}, \theta^c_t)-\text{Reg}_2(\tau, \bm{\theta}^c)\\
&=\sum_{t=1}^{\tau} \left(p(\bm{c}_t)+f_{\theta^c_t}(\bm{x}_t)\right)-\mu\cdot \frac{\tau}{T}+\sum_{t=1}^{\tau}\frac{\mu\cdot g_{i',\theta^c_t}(\bc_t, \bm{x}_t)}{T\cdot\beta_{i'}}-\text{Reg}_2(\tau, \bm{\theta}^c)\\
&\geq \sum_{t=1}^{\tau} \left(p(\bm{c}_t)+f_{\theta^c_t}(\bm{x}_t)\right)+\mu\cdot\frac{T-\tau}{T}-\text{Reg}_2(\tau, \bm{\theta}^c)
\end{aligned}\]
where the last inequality follows from \eqref{eqn:121702}. Then, it holds that
\begin{equation}\label{eqn:121703}
\mathbb{E}_{\bm{\theta}\sim\bm{P}}\left[ \sum_{t=1}^{\tau}\bLInner(\bm{c}_t, \mu\cdot\bm{e}_{i_t}, \theta^c_t) \right]\geq \mathbb{E}_{\bm{\theta}\sim\bm{P}}\left[ \sum_{t=1}^{\tau} \left(p(\bm{c}_t)+f_{\theta^c_t}(\bm{x}_t)\right) \right]+\mu\cdot\frac{T-\tau}{T}-\mathbb{E}_{\bm{\theta}\sim\bm{P}}\left[ \text{Reg}_2(\tau, \bm{\theta}^c) \right].
\end{equation}
Combining \eqref{eqn:121701} and \eqref{eqn:121703}, we have
\[\begin{aligned}
\mathbb{E}_{\bm{\theta}\sim\bm{P}}\left[ \sum_{t=1}^{\tau} \left(p(\bm{c}_t)+f_{\theta^c_t}(\bm{x}_t)\right) \right]\leq& -\mu\cdot\frac{T-\tau}{T}+\tau\cdot \mathbb{E}_{\tbc^*, \theta\sim P}\left[ \bLInner(\tbc^*, \bm{\lambda}^*, \theta) \right]+O(\mathbb{E}_{\bm{\theta}\sim\bm{P}}[W(\bm{\theta})])\\
&+\mathbb{E}_{\bm{\theta}\sim\bm{P}}\left[\text{Reg}_1(\tau, \bm{\theta})\right]+\mathbb{E}_{\bm{\theta}\sim\bm{P}}\left[ \text{Reg}_2(\tau, \bm{\theta}) \right].
\end{aligned}\]
From the boundedness conditions in \Cref{assump:main}, we have
\[
\mathbb{E}_{\tbc^*, \theta\sim P}\left[\bLInner(\tbc^*, \bm{\lambda}^*, \theta)\right]=\frac{1}{T}\cdot\OPTInner\geq -1
\]
which implies that
\[
-\mu\cdot\frac{T-\tau}{T}\leq (T-\tau)\cdot \mathbb{E}_{\tbc^*, \theta\sim P}\left[\bLInner(\tbc^*, \bm{\lambda}^*, \theta)\right]
\]
when $\mu=T$. Therefore, we have
\begin{align}
\mathbb{E}_{\bm{\theta}\sim\bm{P}}\left[ \sum_{t=1}^{\tau} \left(p(\bm{c}_t)+f_{\theta^c_t}(\bm{x}_t)\right) \right]\leq& T\cdot \mathbb{E}_{\tbc^*, \theta\sim P}\left[ \bLInner(\tbc^*, \bm{\lambda}^*, \theta) \right]+O(\mathbb{E}_{\bm{\theta}\sim\bm{P}}[W(\bm{\theta})])+\mathbb{E}_{\bm{\theta}\sim\bm{P}}\left[\text{Reg}_1(T, \bm{\theta})\right] \nonumber\\
&+\mathbb{E}_{\bm{\theta}\sim\bm{P}}\left[ \text{Reg}_2(T, \bm{\theta}) \right]\label{eqn:121704}\\
\leq& \OPT^c+O(\mathbb{E}_{\bm{\theta}\sim\bm{P}}[W(\bm{\theta})])+\mathbb{E}_{\bm{\theta}\sim\bm{P}}\left[\text{Reg}_1(T, \bm{\theta})\right] \nonumber\\
&+\mathbb{E}_{\bm{\theta}\sim\bm{P}}\left[ \text{Reg}_2(T, \bm{\theta}) \right]. \nonumber
\end{align}
where $\OPT^c$ denotes the value of the optimal policy with adversarial corruptions. It is clear to see that $|\OPT^c-\OPT|\leq O(\mathbb{E}_{\bm{\theta}\sim\bm{P}}[W(\bm{\theta})])$.
Using \Cref{thm:OGDregret} and \Cref{thm:Hedgeregret} to bound $\mathbb{E}_{\bm{\theta}\sim\bm{P}}\left[\text{Reg}_1(T, \bm{\theta})\right]$ and $\mathbb{E}_{\bm{\theta}\sim\bm{P}}\left[ \text{Reg}_2(T, \bm{\theta}) \right]$, we have
\[
\mathbb{E}_{\bm{\theta}\sim\bm{P}}\left[ \sum_{t=1}^{\tau} \left(p(\bm{c}_t)+f_{\theta^c_t}(\bm{x}_t)\right) \right]\leq \OPT^c+O((G+F)\cdot\sqrt{T})+O(\sqrt{T\cdot\log m})+O(\mathbb{E}_{\bm{\theta}\sim\bm{P}}[W(\bm{\theta})])
\]
which completes our proof of \eqref{eqn:121801}.
\end{myproof}

\section{Missing Proofs for Appendix \ref{sec:extensions}}\label{sec:pfextensions}

\begin{myproof}[Proof of \Cref{thm:DiscreteC}]
The proof follows a similar procedure as the proof of \Cref{thm:CorruptRegretISP}.
Denote by $\tau$ the time period that \Cref{alg:DAL} is terminated. There must be a constraint $i'\in[m]$ such that
\begin{equation}\label{eqn:012502}
\sum_{t=1}^{\tau} g_{i', \theta^c_t}(\bc_t, \bm{x}_t)\geq T\cdot \beta_{i'}.
\end{equation}
Otherwise, we can assume without loss of generality that there exists a \textit{dummy} constraint $i'$ such that $g_{i',\theta}(\bc_t, \bm{x})=\beta_{i'}=\alpha$, for arbitrary $\alpha\in(0,1)$, for any $\theta$ and $\bc, \bm{x}$. In this case, we can set $\tau=T$.

We denote by $(\bm{\lambda}^*, \tbc^*)$ one \textit{saddle-point} optimal solution to
\begin{equation}\label{eqn:012506}
\max_{\bm{\lambda}\geq0}\min_{\tbc\in\mathcal{C}} \mathbb{E}_{\tbc^*, \theta\sim P^c}\left[\bLInner(\tbc, \bm{\lambda}, \theta)\right]=\min_{\tbc\in\mathcal{C}}\max_{\bm{\lambda}\geq0} \mathbb{E}_{\tbc^*, \theta\sim P^c}\left[\bLInner(\tbc, \bm{\lambda}, \theta)\right]
\end{equation}
where $P^c$ denotes the uniform mixture of the distributions of $\{\theta^c_t\}$ for $t=1$ to $\tau$. We note that since $\tbc$ is a distribution over $K$ discrete points $\{\bm{c}^1,\dots,\bm{c}^K\}$, $\tbc$ can be denoted by $\bm{p}_{\tbc}\in\mathbb{R}^m$ with $0\leq\bm{p}_{\tbc}$ and $\|\bm{p}_{\tbc}\|_1=1$. Then, we have
\[
\mathbb{E}_{\tbc^*, \theta\sim P^c}\left[\bLInner(\tbc, \bm{\lambda}, \theta)\right]=\langle \bm{p}_{\tbc}, \bar{\bm{L}} \rangle
\]
with $\bar{\bm{L}}=( \bLInner(\bc^k, \bm{\lambda}, \theta) )_{k=1}^K\in\mathbb{R}^K$. Therefore, $\mathbb{E}_{\tbc^*, \theta\sim P^c}\left[\bLInner(\tbc, \bm{\lambda}, \theta)\right]$ can be regarded as a convex function over $\tbc$ and is clearly a concave function over $\bm{\lambda}$. The equality \eqref{eqn:012506} follows from the Sion's minimax theorem \citep{sion1958general}.

We have
\[
\sum_{t=1}^{\tau}\bLInner(\bm{c}_t, \mu\cdot\bm{e}_{i_t}, \theta^c_t)\leq \sum_{t=1}^{\tau}\mathbb{E}_{\tbc^*}\left[\bLInner(\tbc^*, \mu\cdot\bm{e}_{i_t}, \theta^c_t)\right]+\text{Reg}_1(\tau, \bm{\theta}^c)
\]
following regret bound of $\ALG_1$. Then, it holds that
\begin{align}
\mathbb{E}_{\bm{\theta}\sim\bm{P}}\left[ \sum_{t=1}^{\tau}\bLInner(\bm{c}_t, \mu\cdot\bm{e}_{i_t}, \theta^c_t) \right]&\leq \mathbb{E}_{\tbc^*, \bm{\theta}\sim\bm{P}}\left[ \sum_{t=1}^{\tau}\bLInner(\tbc^*, \mu\cdot\bm{e}_{i_t}, \theta^c_t) \right]+\mathbb{E}_{\bm{\theta}\sim\bm{P}}\left[\text{Reg}_1(\tau, \bm{\theta}^c)\right] \label{eqn:012503}\\
&\leq \tau\cdot \mathbb{E}_{\tbc^*, \theta\sim P^c}\left[ \bLInner(\tbc^*, \bm{\lambda}^*, \theta) \right]+\mathbb{E}_{\bm{\theta}\sim\bm{P}}\left[\text{Reg}_1(\tau, \bm{\theta}^c)\right] \nonumber\\
&\leq \tau\cdot \mathbb{E}_{\tbc^*, \theta\sim P}\left[ \bLInner(\tbc^*, \bm{\lambda}^*, \theta) \right]+\mathbb{E}_{\bm{\theta}\sim\bm{P}}\left[\text{Reg}_1(\tau, \bm{\theta}^c)\right]+O(\mathbb{E}_{\bm{\theta}\sim\bm{P}}[W(\bm{\theta})]). \nonumber
\end{align}
On the other hand, for any $i\in[m]$, we have
\[
\sum_{t=1}^{\tau}\bLInner(\bm{c}_t, \mu\cdot\bm{e}_{i_t}, \theta^c_t)\geq \sum_{t=1}^{\tau}\bLInner_i(\bm{c}_t, \mu\cdot\bm{e}_{i_t}, \theta^c_t)-\text{Reg}_2(\tau, \bm{\theta}^c)
\]
following the regret bound of $\ALG_2$ (holds for arbitrary $\bm{\lambda}=\mu\cdot\bm{e}_i$). We now set $i=i'$ and we have
\[\begin{aligned}
\sum_{t=1}^{\tau}\bLInner(\bm{c}_t, \mu\cdot\bm{e}_{i_t}, \theta^c_t)&\geq \sum_{t=1}^{\tau}\bLInner_{i'}(\bm{c}_t, \mu\cdot\bm{e}_{i_t}, \theta^c_t)-\text{Reg}_2(\tau, \bm{\theta}^c)\\
&=\sum_{t=1}^{\tau} \left(p(\bm{c}_t)+f_{\theta^c_t}(\bm{x}_t)\right)-\mu\cdot \frac{\tau}{T}+\sum_{t=1}^{\tau}\frac{\mu\cdot g_{i',\theta^c_t}(\bc_t, \bm{x}_t)}{T\cdot\beta_{i'}}-\text{Reg}_2(\tau, \bm{\theta}^c)\\
&\geq \sum_{t=1}^{\tau} \left(p(\bm{c}_t)+f_{\theta^c_t}(\bm{x}_t)\right)+\mu\cdot\frac{T-\tau}{T}-\text{Reg}_2(\tau, \bm{\theta}^c)
\end{aligned}\]
where the last inequality follows from \eqref{eqn:012502}. Then, it holds that
\begin{equation}\label{eqn:012504}
\mathbb{E}_{\bm{\theta}\sim\bm{P}}\left[ \sum_{t=1}^{\tau}\bLInner(\bm{c}_t, \mu\cdot\bm{e}_{i_t}, \theta^c_t) \right]\geq \mathbb{E}_{\bm{\theta}\sim\bm{P}}\left[ \sum_{t=1}^{\tau} \left(p(\bm{c}_t)+f_{\theta^c_t}(\bm{x}_t)\right) \right]+\mu\cdot\frac{T-\tau}{T}-\mathbb{E}_{\bm{\theta}\sim\bm{P}}\left[ \text{Reg}_2(\tau, \bm{\theta}^c) \right].
\end{equation}
Combining \eqref{eqn:012503} and \eqref{eqn:012504}, we have
\[\begin{aligned}
\mathbb{E}_{\bm{\theta}\sim\bm{P}}\left[ \sum_{t=1}^{\tau} \left(p(\bm{c}_t)+f_{\theta^c_t}(\bm{x}_t)\right) \right]\leq& -\mu\cdot\frac{T-\tau}{T}+\tau\cdot \mathbb{E}_{\tbc^*, \theta\sim P}\left[ \bLInner(\tbc^*, \bm{\lambda}^*, \theta) \right]+O(\mathbb{E}_{\bm{\theta}\sim\bm{P}}[W(\bm{\theta})])\\
&+\mathbb{E}_{\bm{\theta}\sim\bm{P}}\left[\text{Reg}_1(\tau, \bm{\theta})\right]+\mathbb{E}_{\bm{\theta}\sim\bm{P}}\left[ \text{Reg}_2(\tau, \bm{\theta}) \right].
\end{aligned}\]
From the boundedness condition c in \Cref{assump:main}, we have
\[
\mathbb{E}_{\tbc^*, \theta\sim P}\left[\bLInner(\tbc^*, \bm{\lambda}^*, \theta)\right]=\frac{1}{T}\cdot\OPTInner\geq -1
\]
which implies that
\[
-\mu\cdot\frac{T-\tau}{T}\leq (T-\tau)\cdot \mathbb{E}_{\tbc^*, \theta\sim P}\left[\bLInner(\tbc^*, \bm{\lambda}^*, \theta)\right]
\]
when $\mu=T$. Therefore, we have
\begin{align}
\mathbb{E}_{\bm{\theta}\sim\bm{P}}\left[ \sum_{t=1}^{\tau} \left(p(\bm{c}_t)+f_{\theta^c_t}(\bm{x}_t)\right) \right]\leq& T\cdot \mathbb{E}_{\tbc^*, \theta\sim P}\left[ \bLInner(\tbc^*, \bm{\lambda}^*, \theta) \right]+O(\mathbb{E}_{\bm{\theta}\sim\bm{P}}[W(\bm{\theta})])+\mathbb{E}_{\bm{\theta}\sim\bm{P}}\left[\text{Reg}_1(T, \bm{\theta})\right] \nonumber\\
&+\mathbb{E}_{\bm{\theta}\sim\bm{P}}\left[ \text{Reg}_2(T, \bm{\theta}) \right]\label{eqn:012505}\\
\leq& \OPT^c+O(\mathbb{E}_{\bm{\theta}\sim\bm{P}}[W(\bm{\theta})])+\mathbb{E}_{\bm{\theta}\sim\bm{P}}\left[\text{Reg}_1(T, \bm{\theta})\right] \nonumber\\
&+\mathbb{E}_{\bm{\theta}\sim\bm{P}}\left[ \text{Reg}_2(T, \bm{\theta}) \right]. \nonumber
\end{align}
where $\OPT^c$ denotes the value of the optimal policy with adversarial corruptions. It is clear to see that $|\OPT^c-\OPT|\leq O(\mathbb{E}_{\bm{\theta}\sim\bm{P}}[W(\bm{\theta})])$.
Using Theorem 3.1 in \cite{auer2002nonstochastic} to bound $\mathbb{E}_{\bm{\theta}\sim\bm{P}}\left[\text{Reg}_1(T, \bm{\theta})\right]$ and \Cref{thm:Hedgeregret} to bound $\mathbb{E}_{\bm{\theta}\sim\bm{P}}\left[ \text{Reg}_2(T, \bm{\theta}) \right]$, we have
\[
\mathbb{E}_{\bm{\theta}\sim\bm{P}}\left[ \sum_{t=1}^{\tau} \left(p(\bm{c}_t)+f_{\theta^c_t}(\bm{x}_t)\right) \right]\leq \OPT^c+O((G+F)\cdot\sqrt{T})+O(\sqrt{T\cdot\log m})+O(\mathbb{E}_{\bm{\theta}\sim\bm{P}}[W(\bm{\theta})])
\]
which completes our proof of \eqref{eqn:012501}.
\end{myproof}

\begin{myproof}[Proof of \Cref{thm:RegretOSP}]
From the formulation \eqref{eqn:LOSPstationary}, the dual variable $\bm{\lambda}$ is scaled by $T$. Therefore, we know that $\mu=\alpha\cdot T$ for some constant $\alpha>0$.

We assume without loss of generality that there always exists $i'\in[m]$ such that
\begin{equation}\label{eqn:121602}
\sum_{t=1}^{T} g_{i', \theta_t}(\bc_t, \bm{x}_t)\leq T\cdot \beta_{i'}.
\end{equation}
In fact, let there be a \textit{dummy} constraint $i'$ such that $g_{i',\theta}(\bc, \bm{x})=\beta_{i'}=\alpha$, for arbitrary $\alpha\in(0,1)$, for any $\theta$ and $\bc, \bm{x}$. Then, \eqref{eqn:121602} holds.

We denote by $(\bm{\lambda}^*, \tbc^*)$ one \textit{saddle-point} optimal solution to
\[
\max_{\bm{\lambda}\geq0}\min_{\tbc\in\mathcal{C}} \mathbb{E}_{\tbc, \theta\sim P}\left[\bLOuter(\tbc, \bm{\lambda}, \theta)\right]=\min_{\tbc\in\mathcal{C}}\max_{\bm{\lambda}\geq0} \mathbb{E}_{\tbc, \theta\sim P}\left[\bLOuter(\tbc, \bm{\lambda}, \theta)\right].
\]
We have
\[
\sum_{t=1}^{T}\bLOuter(\bm{c}_t, \mu\cdot\bm{e}_{i_t}, \theta_t)\leq \sum_{t=1}^{T}\mathbb{E}_{\tbc^*}\left[\bLOuter(\tbc^*, \mu\cdot\bm{e}_{i_t}, \theta_t)\right]+\text{Reg}_1(T, \bm{\theta})
\]
following regret bound of $\ALG_1$. Then, it holds that
\begin{equation}\label{eqn:121601}
\begin{aligned}
\mathbb{E}_{\bm{\theta}\sim\bm{P}}\left[ \sum_{t=1}^{T}\bLOuter(\bm{c}_t, \mu\cdot\bm{e}_{i_t}, \theta_t) \right]&\leq \mathbb{E}_{\tbc^*, \bm{\theta}\sim\bm{P}}\left[ \sum_{t=1}^{T}\bLOuter(\tbc^*, \mu\cdot\bm{e}_{i_t}, \theta_t) \right]+\mathbb{E}_{\bm{\theta}\sim\bm{P}}\left[\text{Reg}_1(T, \bm{\theta})\right]\\
&\leq \tau\cdot \mathbb{E}_{\tbc^*, \theta\sim P}\left[ \bLOuter(\tbc^*, \bm{\lambda}^*, \theta) \right]+\mathbb{E}_{\bm{\theta}\sim\bm{P}}\left[\text{Reg}_1(T, \bm{\theta})\right].
\end{aligned}
\end{equation}
On the other hand, for any $i\in[m]$, we have
\[
\sum_{t=1}^{T}\bLOuter(\bm{c}_t, \mu\cdot\bm{e}_{i_t}, \theta_t)\geq \sum_{t=1}^{T}\bLOuter_i(\bm{c}_t, \mu\cdot\bm{e}_{i_t}, \theta_t)-\text{Reg}_2(T, \bm{\theta})
\]
following the regret bound of $\ALG_2$ in (holds for arbitrary $\bm{\lambda}=\mu\cdot\bm{e}_i$). We now denote by
\[
i^*=\text{argmax}_{i\in[m]}\{ \beta_i-\frac{1}{T}\cdot\sum_{t=1}^{T}g_{i,\theta_t}(\bc_t, \bm{x}_t) \}.
\]
We also denote by
\[
d_T(\mathcal{A}, \bm{\theta})=\max_{i\in[m]}\{ \beta_i-\frac{1}{T}\cdot\sum_{t=1}^{T}g_{i,\theta_t}(\bc_t, \bm{x}_t) \}
\]
as the distance away from the target set $\mathcal{A}$. Following \eqref{eqn:121602}, we always have $d_T(\mathcal{A}, \bm{\theta})\geq0$.
We now set $i=i^*$ and we have
\[\begin{aligned}
\sum_{t=1}^{T}\bLOuter(\bm{c}_t, \mu\cdot\bm{e}_{i_t}, \theta_t)&\geq \sum_{t=1}^{T}\bLInner_{i^*}(\bm{c}_t, \mu\cdot\bm{e}_{i_t}, \theta_t)-\text{Reg}_2(T, \bm{\theta})\\
&=\sum_{t=1}^{T} \left(p(\bm{c}_t)+f_{\theta_t}(\bm{x}_t)\right)+\mu-\sum_{t=1}^{T}\frac{\mu\cdot g_{i^*,\theta_t}(\bc_t, \bm{x}_t)}{T\cdot\beta_{i^*}}-\text{Reg}_2(T, \bm{\theta})\\
&\geq \sum_{t=1}^{T} \left(p(\bm{c}_t)+f_{\theta_t}(\bm{x}_t)\right)+\frac{\mu}{\beta_{i^*}}\cdot d_T(\mathcal{A}, \bm{\theta})-\text{Reg}_2(T, \bm{\theta})
\end{aligned}\]
where the last inequality follows from \eqref{eqn:121602}. Then, it holds that
\begin{equation}\label{eqn:121603}
\begin{aligned}
\mathbb{E}_{\bm{\theta}\sim\bm{P}}\left[ \sum_{t=1}^{T}\bLOuter(\bm{c}_t, \mu\cdot\bm{e}_{i_t}, \theta_t) \right]\geq & \mathbb{E}_{\bm{\theta}\sim\bm{P}}\left[ \sum_{t=1}^{T} \left(p(\bm{c}_t)+f_{\theta_t}(\bm{x}_t)\right) \right]\\
&+\frac{\mu}{\beta_{\max}}\cdot \mathbb{E}_{\bm{\theta}\sim\bm{P}} [d_T(\mathcal{A}, \bm{\theta})]-\mathbb{E}_{\bm{\theta}\sim\bm{P}}\left[ \text{Reg}_2(T, \bm{\theta}) \right].
\end{aligned}
\end{equation}
where $\beta_{\max}=\max_{i\in[m]}\{\beta_i\}$.
Combining \eqref{eqn:121601} and \eqref{eqn:121603}, we have
\begin{equation}\label{eqn:121604}
\begin{aligned}
\mathbb{E}_{\bm{\theta}\sim\bm{P}}\left[ \sum_{t=1}^{T} \left(p(\bm{c}_t)+f_{\theta_t}(\bm{x}_t)\right) \right]+\frac{\mu}{\beta_{\max}}\cdot \mathbb{E}_{\bm{\theta}\sim\bm{P}} [d_T(\mathcal{A}, \bm{\theta})]\leq& T\cdot \mathbb{E}_{\theta\sim P}\left[ \bLOuter(\bm{c}^*, \bm{\lambda}^*, \theta) \right]\\
&+\mathbb{E}_{\bm{\theta}\sim\bm{P}}\left[\text{Reg}_1(T, \bm{\theta})\right]+\mathbb{E}_{\bm{\theta}\sim\bm{P}}\left[ \text{Reg}_2(T, \bm{\theta}) \right].
\end{aligned}
\end{equation}
From the non-negativity of $d_T(\mathcal{A}, \bm{\theta})$, we have
\[
\mathbb{E}_{\bm{\theta}\sim\bm{P}}\left[ \sum_{t=1}^{T} \left(p(\bm{c}_t)+f_{\theta_t}(\bm{x}_t)\right) \right]\leq T\cdot \mathbb{E}_{\theta\sim P}\left[ \bLOuter(\bm{c}^*, \bm{\lambda}^*, \theta) \right]+\mathbb{E}_{\bm{\theta}\sim\bm{P}}\left[\text{Reg}_1(T, \bm{\theta})\right]+\mathbb{E}_{\bm{\theta}\sim\bm{P}}\left[ \text{Reg}_2(T, \bm{\theta}) \right].
\]
Using \Cref{thm:OGDregret} and \Cref{thm:Hedgeregret} to bound $\mathbb{E}_{\bm{\theta}\sim\bm{P}}\left[\text{Reg}_1(T, \bm{\theta})\right]$ and $\mathbb{E}_{\bm{\theta}\sim\bm{P}}\left[ \text{Reg}_2(T, \bm{\theta}) \right]$, we have
\[
\mathbb{E}_{\bm{\theta}\sim\bm{P}}\left[ \sum_{t=1}^{T} \left(p(\bm{c}_t)+f_{\theta_t}(\bm{x}_t)\right) \right]\leq T\cdot \mathbb{E}_{\theta\sim P}\left[ \bLOuter(\bm{c}^*, \bm{\lambda}^*, \theta) \right]+O((G+F)\cdot\sqrt{T})+O(\sqrt{T\cdot\log m})
\]
which completes our proof of \eqref{eqn:121501}.

We now prove \eqref{eqn:121502}. We note that $(\bm{c}_t, \bm{x}_t)_{t=1}^T$ defines a feasible solution to
\begin{align}
 \OPT^{\delta}= \min & \sum_{t=1}^{T}\mathbb{E}_{\bm{c}_t, \bm{x}_t, \theta_t}\left[p(\bm{c}_t)+f_{\theta_t}(\bm{x}_t)\right] \label{lp:OuterA}\\
  \mbox{s.t.} & \frac{1}{T}\cdot\sum_{t=1}^{T}\mathbb{E}_{\bc_t, \bm{x}_t, \theta_t}[\bm{g}_{\theta_t}(\bc_t, \bm{x}_t)]\geq\bm{\beta}-\delta \nonumber\\
  & \bm{x}_t\in\mathcal{K}(\theta_t, \bm{c}_t), \bm{c}_t\in\mathcal{C}, \forall t.\nonumber
\end{align}
with $\delta=\mathbb{E}_{\bm{\theta}\sim\bm{P}}[d_T(\mathcal{A}, \bm{\theta})]$. If we regard $\OPT^{\delta}$ as a function over $\delta$, then $\OPT^{\delta}$ is clearly a convex function over $\delta$, where the proof follows the same spirit as the proof of \Cref{lem:ConvexL}. Moreover, note that
\[
\frac{d\OPT^{\delta=0}}{d\delta}=\|\bm{\lambda^*}\|_1\leq\mu.
\]
We have
\[
\OPTOuter=\OPT^{0}\leq \OPT^{\delta}+  \mu\cdot \delta
\]
for any $\delta\geq0$. Therefore, it holds that
\begin{equation}\label{eqn:121605}
\begin{aligned}
T\cdot \mathbb{E}_{\theta\sim P}\left[ \bLOuter(\bm{c}^*, \bm{\lambda}^*, \theta) \right]&=\OPTOuter \leq \OPT^{\mathbb{E}_{\bm{\theta}\sim\bm{P}}[d_T(\mathcal{A},\bm{\theta})]}+ \mu\cdot \mathbb{E}_{\bm{\theta}\sim\bm{P}}[d_T(\mathcal{A},\bm{\theta})]\\
&\leq \sum_{t=1}^{T} \mathbb{E}_{\bm{\theta}\sim\bm{P}}\left[p(\bm{c}_t)+f_{\theta_t}(\bm{x}_t)+\mu\cdot d_T(\mathcal{A},\bm{\theta})\right]
\end{aligned}
\end{equation}
Plugging \eqref{eqn:121605} into \eqref{eqn:121604}, we have
\[
\mu\cdot\left(\frac{1}{\beta_{\max}}-1\right)\cdot \mathbb{E}_{\bm{\theta}\sim\bm{P}}[ d_T(\mathcal{A},\bm{\theta})]\leq \mathbb{E}_{\bm{\theta}\sim\bm{P}}\left[\text{Reg}_1(T, \bm{\theta})\right]+\mathbb{E}_{\bm{\theta}\sim\bm{P}}\left[ \text{Reg}_2(T, \bm{\theta}) \right].
\]
Our proof of \eqref{eqn:121502} is completed by using \Cref{thm:OGDregret} and \Cref{thm:Hedgeregret} to bound $\mathbb{E}_{\bm{\theta}\sim\bm{P}}\left[\text{Reg}_1(T, \bm{\theta})\right]$ and $\mathbb{E}_{\bm{\theta}\sim\bm{P}}\left[ \text{Reg}_2(T, \bm{\theta}) \right]$.
\end{myproof}

\end{APPENDICES}

\end{document}